\documentclass[10pt,twocolumn,letterpaper]{article}

\usepackage[pagenumbers]{cvpr} 

\usepackage{booktabs}
\usepackage{multirow}
\usepackage{xspace}
\usepackage{subcaption}
\usepackage{algorithm}
\usepackage{algpseudocode}
\usepackage{listings}
\usepackage[most]{tcolorbox}
\definecolor{inframe}{HTML}{A3A9B1}
\definecolor{inback}{HTML}{F7F7F8}
\definecolor{outframe}{HTML}{8AA3C2}
\definecolor{outback}{HTML}{F2F6FB}
\definecolor{boxhead}{HTML}{2E4A6B}
\lstdefinestyle{prompt}{basicstyle=\fontencoding{T1}\fontfamily{phv}\fontsize{7.5}{9.8}\selectfont\color{black!88}, columns=fullflexible, keepspaces=true,
  breaklines=true, breakatwhitespace=true, breakindent=0pt, showstringspaces=false, aboveskip=0pt, belowskip=0pt, literate={-}{-}1,
  morecomment=[l][\bfseries\color{boxhead}]{\#}}
\tcbset{promptbase/.style={enhanced, listing only, listing options={style=prompt}, arc=3pt, boxrule=0.5pt, left=6pt, right=6pt, top=3pt, bottom=3pt,
  fonttitle=\fontencoding{T1}\fontfamily{phv}\fontseries{b}\fontsize{7.5}{9}\selectfont, toptitle=1.5pt, bottomtitle=1.5pt, titlerule=0.3pt, before skip=3pt, after skip=3pt}}
\newtcblisting{promptbox}[1]{promptbase, colframe=inframe, colback=inback, colbacktitle=inframe!22, coltitle=black!75, title={#1}}
\newtcblisting{agentbox}{promptbase, colframe=outframe, colback=outback, colbacktitle=outframe!28, coltitle=boxhead, title={Agent: tool call}}
\newtcblisting{resultbox}{promptbase, colframe=inframe, colback=inback, colbacktitle=inframe!22, coltitle=black!75, title={Result returned to the agent}}

\definecolor{cvprblue}{rgb}{0.21,0.49,0.74}
\usepackage[pagebackref,breaklinks,colorlinks,allcolors=cvprblue]{hyperref}

\title{Inhabitant: An Agent Framework for Long-Horizon Human-Scene Interaction\\with Spatial Control and Quantitative Feedback}

\author{Yao Zhang\textsuperscript{*} \qquad Zhuchenyang Liu\textsuperscript{*} \qquad Yu Xiao\\
Aalto University\\
{\small \textsuperscript{*}Equal contribution}\\
{\small Project page: \url{https://ryenhails.github.io/Inhabitant/}}
}

\begin{document}
\maketitle

\begin{abstract}
Recent work on human-scene interaction (HSI) generation moves from single interactions to long-horizon interaction, where a language-model agent plans and executes a sequence of interactions in a 3D scene under one instruction.
However, existing agents struggle in challenging everyday scenarios (e.g., picking up a toy and throwing it into a basket) because they lack spatial control over the motion and quantitative feedback on its execution, and they often fail to complete the interaction sequence. To address these gaps, we present Inhabitant, a training-free agent framework in which the agent specifies each motion through positions, paths and poses, measures how the executed interaction deviates from its intent, and revises the control specification accordingly.
For an extensive evaluation in such scenarios, we build Inhabit, a benchmark of 75 household tasks in 31 furniture layouts of an apartment, in which every task chains one to three goals on movable objects and is carried out by the whole body, excluding fingers.
On Inhabit, Inhabitant completes 98.7\% of the tasks with Claude Opus 5.5, and 93.3\% and 82.7\% with Claude Sonnet 5.5 and Gemini 4 Argon, respectively. Ablations confirm that without spatial control the agent fails most tasks, and without quantitative feedback its motion becomes less precise.
Code, the benchmark and a video are available on the project page.
\end{abstract}

\section{Introduction}
\label{sec:intro}

Human-scene interaction (HSI) generation, which synthesizes how a person moves through a scene and acts on its objects, is a core problem for computer graphics and embodied AI~\cite{starke2019neural,hassan2021stochastic,hassan2023synthesizing,puig2024habitat}. Early work typically focuses on isolated, short-horizon interactions: a single action involving one target object, such as sitting down on a chair, standing up from a chair, or picking up an object~\cite{zhang2022couch,wang2022humanise,zhao2023synthesizing,li2024controllable,jiang2024scaling,pan2025tokenhsi}.
Recent work has shifted towards a more challenging setting of long-horizon interaction, where a language-model agent plans and executes a sequence of interactions in a 3D scene under one instruction, for example, ``put the cube in the basket, kick the ball under the table and put the cushion back on the sofa''. These scenes are often simulated environments that respond physically to the body~\cite{xiao2024unified,wu2025human,wang2025sims,jian2026endowing,bao2026generative,zhang2026interactagent,siyao2026humanclaw}.

\begin{figure*}[t]
\centering
\includegraphics[width=\linewidth]{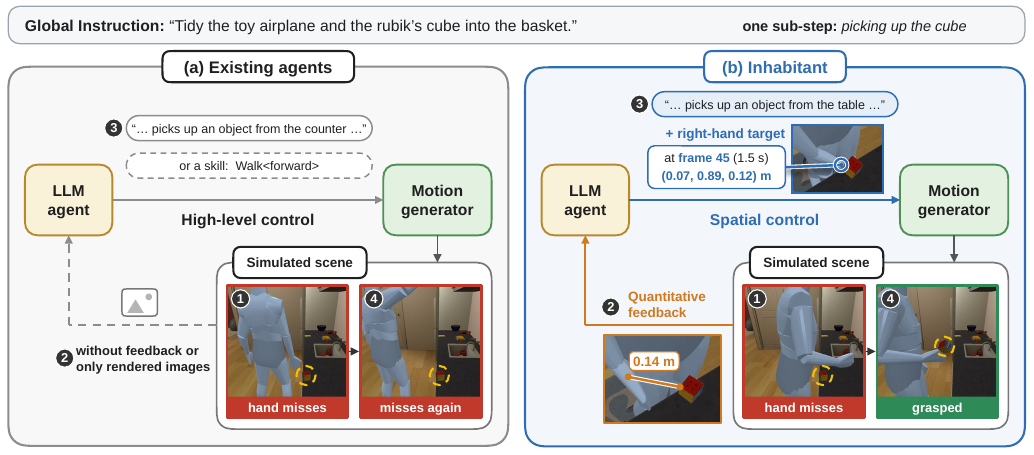}
\caption{One step of a household task: picking up a cube from the counter. (a) Existing agents control the motion through a text prompt or a skill and receive no feedback or only rendered frames; after the hand misses the cube, the agent can only reword the prompt, and the hand misses again. (b) Inhabitant measures how far the hand stopped from the cube (quantitative feedback), fixes where the right hand must be at a given frame (spatial control), and grasps the cube on the next try. The frames in (a) come from our agent restricted to text prompts.}
\label{fig:teaser}
\end{figure*}

However, existing agents often fail to complete such a sequence when every interaction must be executed through full-body motion (Fig.~\ref{fig:teaser}a).
In the example above, the agent first tries to pick up the cube, but the hand does not reach close enough to grasp it and closes a few centimetres away. The agent cannot correct this failure: a text prompt cannot specify the required hand position precisely, and rendered frames do not quantify the spatial error.
We investigate such failures by adapting three representative agents to these scenarios (Sec.~\ref{sec:main}) and attribute them to two causes, both concerning the interface between the agent and the generated interaction.
First, the \textbf{control} is mostly high-level: the agent conditions the human motion only on a skill with one parameter~\cite{siyao2026humanclaw}, a text description of each step~\cite{wu2025motion,zhang2026interactagent}, or a few contact or goal positions~\cite{xiao2024unified,jian2026endowing}. As a result, it cannot specify the motion at the level of precise positions, paths and poses.
Second, the \textbf{feedback} is absent or qualitative: most agents plan the interaction once and never observe its outcome~\cite{xiao2024unified,wu2025human,wang2025sims}, including the few that write poses for a motion generator~\cite{qu2024gpt,bao2026generative}. Other agents observe only rendered frames~\cite{siyao2026humanclaw,zhang2026interactagent}, which may show that an interaction failed but cannot indicate precisely how far the executed interaction deviates from its intent.

We present \textbf{Inhabitant}, a training-free agent framework that turns long-horizon human-scene interaction into a closed-loop spatial control problem (Fig.~\ref{fig:teaser}b). The key novelty is to expose two interfaces that existing agents lack: \emph{spatially grounded control} over generated whole-body motion and \emph{quantitative feedback} on its physical execution.
For control, the agent directly writes the control specification of a pretrained motion generator, Kimodo~\cite{rempe2026kimodo}, and specifies the motion by positions, paths and poses rather than only high-level skills or text prompts.
For feedback, each motion is executed in a scene simulator with half-physics~\cite{siyao2025half};  the agent then queries measurements of the body and the scene, such as positions, distances and contacts, to quantify how the executed interaction deviates from its intent. It revises the control specification using the measured deviation.
The two interfaces close a loop: the agent proposes a motion, executes it, measures the outcome, revises the specification and commits the result as a short motion segment only when the measurements match its intent. By composing such verified segments, Inhabitant builds a complete interaction sequence segment by segment.

For an extensive evaluation in such challenging everyday scenarios, we build \textbf{Inhabit} (Fig.~\ref{fig:bench}), a new benchmark of 75 household tasks in 31 furniture layouts of the ReplicaCAD apartment~\cite{szot2021habitat,yang2025embodiedbench}, simulated with half-physics~\cite{siyao2025half} in line with HumanCLAW~\cite{siyao2026humanclaw}.
Each task states one to three goals, such as placing an object, throwing it or moving it with a foot; ten tasks additionally combine a goal with a body posture, such as sitting, lying or kneeling.
The goals are designed to require coordinated use of the hands, the feet and whole-body posture.
Inhabit differs from previous benchmarks in two respects. First,
HSI benchmarks typically evaluate a single interaction with a static target at a time, such as walking to a piece of furniture and sitting on it~\cite{wang2022humanise,siyao2026humanclaw}. In contrast, Inhabit chains multiple interactions with movable objects under a single instruction, so the outcome of one interaction, including its errors, becomes the initial condition for the next. Second,
embodied household benchmarks also contain long-horizon tasks, but agents usually solve them by calling predefined skills such as navigating, picking and placing~\cite{chang2025partnr,yang2025embodiedbench}. In Inhabit, every step must instead be realized through generated whole-body motion, excluding fingers, so the benchmark tests whether the body can reach, hold and move an object through physically executed motion.
A success rule judges the physical outcome, while collisions, disturbed objects and foot skate measure the quality of the motion.

On Inhabit, Inhabitant completes 74 of the 75 tasks (98.7\%) with Claude Opus 5.5 as the agent, and 93.3\% and 82.7\% of them with Claude Sonnet 5.5 and Gemini 4 Argon, respectively. By comparison, the best of three existing agents adapted to Inhabit completes only 26.7\% with Claude Opus 5.5.
Ablations show that without spatial control the agent fails most tasks despite retaining all feedback, and without quantitative feedback its motion becomes less precise and the agent less reliably recognizes when a task has been completed.



\section{Related Work}
\label{sec:related}

\paragraph{Motion for human-scene interaction.}
Learned HSI methods synthesize interactions in 3D scenes from interaction data, either kinematically~\cite{starke2019neural,hassan2021stochastic,zhang2022wanderings,zhang2022couch,wang2022humanise,zhao2023synthesizing,li2024controllable,yi2024generating,wang2024move,jiang2024autonomous,jiang2024scaling,wang2025hsi} or with physics simulation~\cite{hassan2023synthesizing,wang2023physhoi,wang2025skillmimic,xu2025intermimic,tessler2024maskedmimic,wu2025uniphys,pan2025tokenhsi,xu2024humanvla,zhang2025lhm}. 
These methods are effective for the interaction types and object categories represented in their training data, but are less suited to open-ended long-horizon tasks in which an agent must repeatedly adapt motion to the evolving physical state of a scene. A complementary line of work trains general motion generators without a scene, but allows users to specify spatial constraints in addition to text, such as target positions, paths or poses. Examples include OmniControl~\cite{xie2024omnicontrol}, Kimodo~\cite{rempe2026kimodo}, and related controllable motion generators~\cite{karunratanakul2023guided,wan2024tlcontrol,cohan2024flexible,karunratanakul2024optimizing,pinyoanuntapong2025maskcontrol,zhao2025dartcontrol,li2025genmo,petrovich2024multi,barquero2024seamless}.
Recent systems also let a language model write such constraints, usually once, for a motion generator~\cite{goel2024iterative,liu2024programmable,zheng2026re}.
Inhabitant builds on this capability without training, but writes the constraints repeatedly and revises them from measurements of the executed motion.

\paragraph{Language agents for human-scene interaction.}
Agents that direct a body in a scene differ in what the language model writes: a plan for trained controllers or expert models~\cite{wu2025human,wang2025sims,deng2026human}, such as UniHSI~\cite{xiao2024unified} and World2HOSI~\cite{li2026world2hosi}; atomic skills, such as HumanCLAW~\cite{siyao2026humanclaw}; text for a text-to-motion model, such as Motion-Agent~\cite{wu2025motion} and InteractAgent~\cite{zhang2026interactagent}; or spatial targets for a motion generator~\cite{zhang2026interpose}, such as GPT-Connect~\cite{qu2024gpt}, BiBo~\cite{jian2026endowing} and GEMS~\cite{bao2026generative}.
Among them, InteractAgent is closest in spirit because it revises motion after observing rendered outcomes, but the revision is made through text and relies on qualitative visual feedback. Inhabitant instead revises spatial constraints from metric measurements of the execution.

\paragraph{Embodied household tasks.}
Household benchmarks pose long-horizon tasks with movable objects, such as PARTNR~\cite{chang2025partnr} and EmbodiedBench~\cite{yang2025embodiedbench}, which build on earlier simulated household environments~\cite{puig2018virtualhome,savva2019habitat,shridhar2020alfred,szot2021habitat,li2023behavior,puig2024habitat,li2024embodied,shukla2025maniskill}. Language agents typically solve these tasks in a perception-action loop with feedback from the environment~\cite{yao2022react,huang2022inner,liang2023code,huang2023voxposer,huang2024rekep,rana2023sayplan}. The closest case is FAEA~\cite{tsui2026demonstration}, which refines a manipulation program over repeated attempts in simulation. However, these systems generally act through skills, primitives or end-effector commands, including for the humanoid avatars of Habitat 3.0.
Inhabit keeps the long-horizon household setting but requires every step to be realized by generated whole-body motion.

\section{Inhabitant}
\label{sec:method}

Inhabitant is a training-free agent framework with three components.
The agent is connected to the other two components by two interfaces, which are the core of the framework: \emph{spatial control}, through which it specifies the motion for the generator by positions, paths and poses, and \emph{quantitative feedback}, through which the simulator reports the executed result as measurements.
We first state the problem and then describe the components one by one.

\begin{figure*}[t]
\centering
\includegraphics[width=\linewidth]{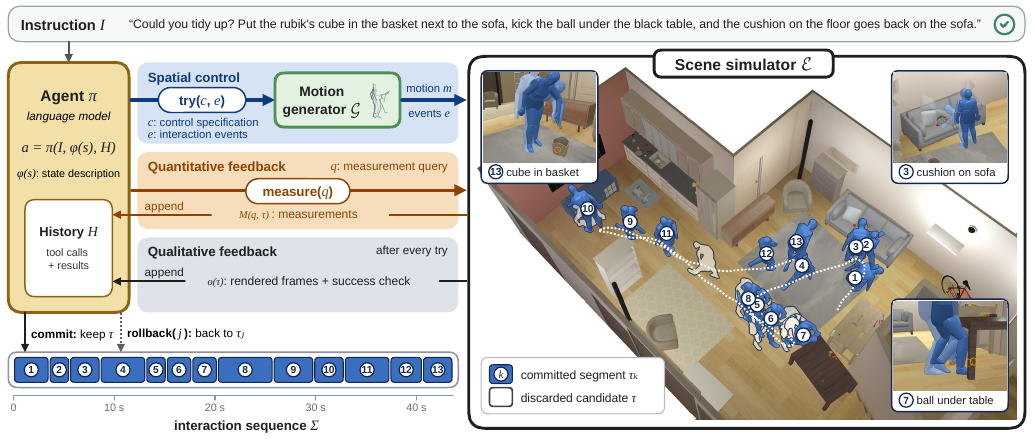}
\caption{Overview of Inhabitant. Left: the agent $\pi$ controls a pretrained motion generator $\mathcal{G}$ by writing a control specification $c$ (spatial control) and interaction events $e$. The scene simulator $\mathcal{E}$ executes the motion, and the agent queries measurements $M(q,\tau)$ of the candidate segment $\tau$ (quantitative feedback) before it tries again, commits or rolls back (Sec.~\ref{sec:llm}). Right: one episode with Claude Opus 5.5, 13 committed segments out of 37 tries. Solid bodies: ends of committed segments, numbered in order; hollow bodies: discarded candidates; insets: three of the interactions; check mark: the success rule is satisfied.}
\label{fig:agent}
\end{figure*}

\subsection{Problem Formulation}
\label{sec:formulation}

An \emph{interaction sequence} is a fixed-frame-rate sequence of states $\Sigma = (s_0, \dots, s_N)$, where each state $s_t$ comprises the poses of the body, all objects and the articulated furniture at frame $t$. Here, $N$ denotes the number of frames.
The agent does not act frame by frame but through \emph{segments}: a segment $\tau_k = (s_{t_{k-1}}, \dots, s_{t_k})$ is a contiguous, few-second portion of the sequence that is generated and executed as a unit.
The full sequence is the concatenation of $K$ segments, $\Sigma = (\tau_1, \dots, \tau_K)$.

The input is a natural-language instruction $I$ and the initial state $s_0$.
Task completion is evaluated by a success rule $G$, which checks the states of the sequence and returns 1 for success and 0 otherwise.
The rule is task-specific, defined together with its instruction (Sec.~\ref{sec:tasks}), and hidden from the agent.
The objective is to produce a successful interaction sequence $\Sigma$ such that $G(\Sigma) = 1$.

\subsection{Agent}
\label{sec:llm}
\label{sec:loop}

The agent $\pi$ is a language model that builds the interaction sequence segment by segment. The prompt of the agent contains the instruction $I$, the state description $\phi(s_0)$ of the initial state, the documentation of the generator, the dimensions of the simulated body, and the request for person-like motion that neither collides with the scene nor slides its feet (Supp.~\ref{sec:supp_prompts}).
At each step it issues one tool call $a$ ~\cite{yao2022react},
\begin{equation}
\begin{aligned}
a = \pi\big(I, \phi(s), H\big) \in \{&\textsc{try}(c, e),\, \textsc{measure}(q),\\
&\textsc{commit},\, \textsc{rollback}(j)\},
\end{aligned}
\label{eq:control}
\end{equation}
where the \emph{state description} $\phi(s)$ lists, in text, the positions and sizes of the body and of the objects in the state $s$ where the preceding segment ended. The agent can re-read the  $\phi(s)$ at any time. $H$ stores previous tool calls and their results.
The agent chooses both the call type and its arguments. The framework prescribes neither the order of the calls nor their content. Next we define the four call types.
\begin{itemize}[leftmargin=*,nosep]
\item $\textsc{try}(c, e)$ requests a motion for the next segment. The control specification $c$ states what the body should do, for how long, and where specified joints should be at specified frames; This is how the agent exerts \emph{spatial control} (Sec.~\ref{sec:request}). The interaction events $e$ state which hand grasps or releases which object, and when. The simulator executes the generated motion from the state at the end of the last segment, and the call returns qualitative feedback $o(\tau)$ on the resulting \emph{candidate segment} $\tau$, i.e., the sequence of states that the execution produces (Sec.~\ref{sec:measure}). A candidate segment is not added to the sequence until it is committed; a subsequent $\textsc{try}$ replaces it, allowing the agent to revise a motion by trying again from the same state.
\item $\textsc{measure}(q)$ asks the simulator for a measurement and returns its value $M(q, \tau)$. The agent chooses the query $q$: what quantity to measure, between which entities, and at which frame or interval. For example, it may ask for the distance between a hand and an object at a given frame of the last try. These measurements are the \emph{quantitative feedback}, and the agent receives only the measurements it asks for. The available queries, which may refer to an earlier segment or to the current state, are given in Sec.~\ref{sec:measure}.
\item $\textsc{commit}$ appends the candidate segment of the last try to the interaction sequence, $\tau_{k+1} = \tau$, and returns the state description of the new state.
\item $\textsc{rollback}(j)$ restores the state $s_{t_j}$ at the end of an earlier segment $j$, discards all later committed segments, and returns the state description of the restored state.
\end{itemize}

These calls build the interaction sequence in a loop; pseudocode is given in the supplement (Algorithm~\ref{alg:loop}).
The loop ends when the success rule is satisfied after a commit, when the agent stops issuing calls, or when the budget of calls to the language model is used up. The sequence $\Sigma$ built up to that point is returned.


\subsection{Motion Generator: Spatial Control}
\label{sec:request}

The motion generator $\mathcal{G}$ provides the spatial control interface to the agent.
We adopt Kimodo~\cite{rempe2026kimodo}, a pretrained generator of whole-body motion conditioned on text and spatial constraints. For each $\textsc{try}$ call, 
Kimodo converts the control specification into a motion
\begin{equation}
m = \mathcal{G}(c;\, h),
\label{eq:generate}
\end{equation}
where $m$ is a sequence of whole-body poses at the frame rate of the sequence, conditioned on the motion $h$ at the end of the preceding segment.

The control specification $c$ 
is written by the agent in the input format of Kimodo, which is documented in its prompt.
It has three parts.
First, a timeline of text prompts with their durations, generated one after another.
Second, a set of spatial constraints $\mathcal{K}$, through which the agent exerts \emph{spatial control}: each fixes a set of specified joints at a specified frame to a target position or pose.
Kimodo supports three types of constraints: \emph{root waypoints}, which fix the position of the root on the ground and optionally its heading, \emph{end-effector targets}, which fix a hand or a foot, and \emph{full-body keyframes}, which fix all joints.
Third, the sampling parameters, namely the random seed and the guidance weights for text and constraints.
The agent uses no intermediate representation between the agent and Kimodo.
Since a full-body keyframe is written as joint rotations, the agent can ask by forward kinematics where the joints of a pose would be, for example a hand, before it uses the pose in a try.
When $\mathcal{K}$ is not available, the agent can only choose the text prompts, their durations and the seed.

The condition $h$ in Eq.~\eqref{eq:generate} is the final frames of the executed body in the preceding segment, converted to the body of the generator and imposed as a full-body constraint before the first frame of the new motion.

The body of the generator and the simulated body differ in their bone lengths.
Before execution, a map $\rho$ transfers the generated motion to the simulated skeleton while keeping the height of the feet above the floor, and then solves inverse kinematics for the joint rotations.
The prompt states the dimensions of the simulated body, so that targets are specified for the body that is executed.

\subsection{Scene Simulator: Feedback}
\label{sec:measure}

The scene simulator $\mathcal{E}$ executes each try and returns qualitative and quantitative feedback on it.

For a try of segment $k{+}1$, the simulator starts from the state at the end of segment $k$ and executes the retargeted motion together with its events.
\begin{equation}
\tau = \mathcal{E}\big(s_{t_k},\, \rho(m),\, e\big).
\label{eq:exec}
\end{equation}
The result is a candidate segment $\tau$.
During execution, the body may deviate from the generated motion $m$ because it is blocked by the scene, while objects move only through contact with the body.

As in household benchmarks~\cite{szot2021habitat,yang2025embodiedbench}, grasping is abstracted by a rule on the distance between the palm and the object (Sec.~\ref{sec:setting}).
The set $e$ contains grasps and releases events.
A grasp event specifies a hand, an object and a time interval, and attaches the object to the hand if the grasp rule is met at any frame of the interval.
A release event specifies a hand and a frame, and detaches the object held by that hand at that frame.

After each try, the agent automatically receives qualitative feedback $o(\tau)$ on the candidate segment: the success check, which says whether the success rule $G$ is satisfied, and a set of rendered frames (Tab.~\ref{tab:feedback_kinds}).

Quantitative feedback is returned only when the agent issues a $\textsc{measure}(q)$ call. Here, $q$ is a measurement query: it specifies the quantity to measure, the entities involved, and the frame or interval of the candidate segment to which the measurement applies.
The measurement operator returns the corresponding numeric value $M(q,\tau)$.
A measurement is a \emph{body measurement} if its query involves a body part, and a \emph{scene measurement} otherwise.
Tab.~\ref{tab:feedback_kinds} lists the available measurement types.
The measurements use the same coordinates as the control specification.
The agent can therefore converts a measurement into a constraint directly, for example by moving a hand target by the measured distance to an object, or by reusing an executed pose as a keyframe.

\begin{table}[t]
\centering
\small
\setlength{\tabcolsep}{4pt}
\begin{tabular}{@{}lp{0.74\linewidth}@{}}
\toprule
\multicolumn{2}{@{}l}{\emph{Qualitative feedback} $o(\tau)$, after every try}\\
\midrule
Check & whether the success rule $G$ is satisfied by the sequence extended by the candidate segment\\
Frames & eight frames of the candidate segment, each from two third-person cameras: one follows the body, one views the body and the target of the motion from the side\\
\midrule
\multicolumn{2}{@{}l}{\emph{Quantitative feedback} $M(q, \tau)$, on a query}\\
\midrule
Body & (i) position of a body part; (ii) distance from a body part to an object, a surface or a handle; (iii) contacts of the body; (iv) pose that the body executed at a frame\\
Scene & (v) position of an object; (vi) distance between two objects or between an object and a region; (vii) extent of an object or a piece of furniture\\
\bottomrule
\end{tabular}
\caption{The four kinds of feedback that the agent receives on a candidate segment.}
\label{tab:feedback_kinds}
\end{table}


\section{Inhabit}
\label{sec:bench}
\label{sec:protocol}

\begin{figure*}[t]
\centering
\includegraphics[width=\linewidth]{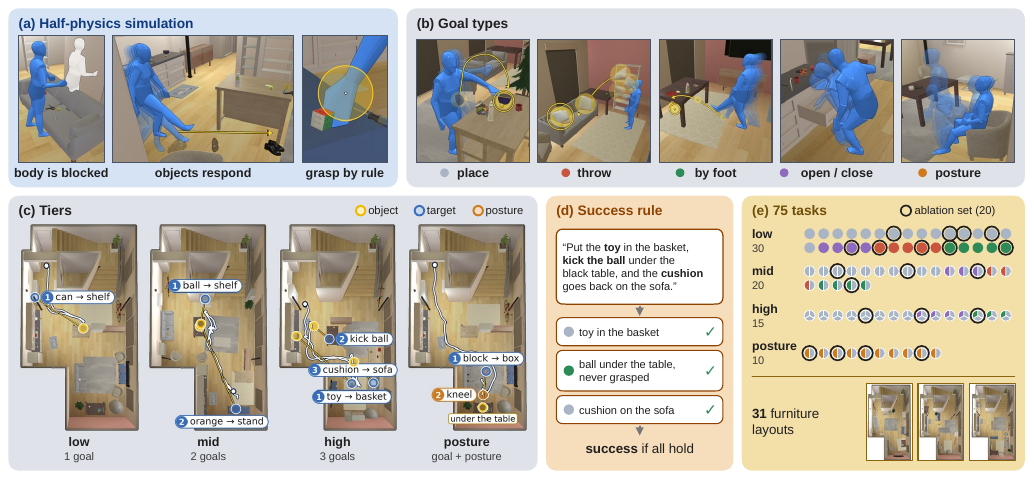}
\caption{Overview of \textbf{Inhabit}. 
(a) Half-physics simulation: the hollow body shows the generated motion before execution, and the solid body shows the executed motion after scene interaction. 
(b) The five goal types, illustrated on segments from an Inhabitant episode. 
(c) Example tasks from each tier in the apartment, showing the body path, each object's start position and its target. 
(d) Success rule for the high-tier task in (c), with one condition per goal. 
(e) The full set of 75 tasks: each circle represents one task, with one sector per goal colored as in (b), alongside three of the 31 furniture layouts.}
\label{fig:bench}
\end{figure*}

\label{sec:setting}
\label{sec:exec}
Inhabit is built on the 31 furniture layouts of the ReplicaCAD apartment used in EB-Habitat~\cite{yang2025embodiedbench,szot2021habitat}.
Its scene simulator $\mathcal{E}$ uses half-physics ~\cite{siyao2025half,siyao2026humanclaw} (Fig.~\ref{fig:bench}a): the body follows the generated motion kinematically but is stopped by furniture and walls, objects respond to the body under rigid-body physics, and grasping and releasing are decided by rules (Supp.~\ref{sec:supp_sim}).

\label{sec:tasks}
Inhabit contains 75 household tasks.
A task $(I, s_0, G)$ consists of an instruction, an initial state and a success rule (Sec.~\ref{sec:formulation}).
Each instruction states one or more goals from five types (Fig.~\ref{fig:bench}b): placing an object, throwing it from at least 1\,m away, moving it with a foot without grasping it, opening or closing a drawer or the fridge, and taking a posture.
The tasks are organized into four tiers (Fig.~\ref{fig:bench}c,\,e): 30 low tasks with one goal, 20 mid tasks with two goals, 15 high tasks with three goals, and 10 posture tasks that combine a goal with a posture.

Each goal gives one condition on the states of the interaction sequence, and the success rule $G$ of a task is the conjunction of its conditions (Fig.~\ref{fig:bench}d).
All conditions are listed in Supp.~\ref{sec:supp_tasks}.

An \emph{episode} is one run of an agent framework on one of these tasks. It ends as described in Sec.~\ref{sec:loop}, with a budget of 150 calls to the language model, and counts as a success only if the final returned sequence satisfies the success rule.

\label{sec:metrics}
On the interaction sequence that an episode returns, we report eight metrics in three groups (Supp.~\ref{sec:supp_metrics}).
For task completion, \emph{Success} is the fraction of episodes whose final state satisfies the success rule, and \emph{Sub-goal} is the mean fraction of success-rule conditions satisfied at the end. 
For motion quality, \emph{Collisions} counts body contacts with furniture and walls, excluding feet on the floor and hands on a grasped object; \emph{Disturbed} counts non-target objects that end more than 5\,cm from their initial positions; and \emph{Skate} is the foot-skate ratio of the sequence~\cite{karunratanakul2023guided}.
For cost, \emph{Tries} counts the motions generated and executed in an episode, \emph{Calls} counts the calls to the language model, and \emph{Tokens} reports the size of the history at the end of the episode.

\section{Experiments}
\label{sec:exp}


\begin{table*}[t]
\centering
\small
\setlength{\tabcolsep}{4.5pt}
\resizebox{\linewidth}{!}{%
\begin{tabular}{llcccccccccccc}
\toprule
& & \multicolumn{6}{c}{Success rate (\%) $\uparrow$} & \multicolumn{3}{c}{Motion quality $\downarrow$} & \multicolumn{3}{c}{Cost $\downarrow$}\\
\cmidrule(lr){3-8}\cmidrule(lr){9-11}\cmidrule(lr){12-14}
Method & Language model & All & low & mid & high & posture & Sub-goal & Collisions & Disturbed & Skate (\%) & Tries & Calls & Tokens\\
\midrule
InteractAgent$^\dagger$~\cite{zhang2026interactagent} & Claude Opus 5.5 & 26.7 & 53.3 & 10.0 & 0.0 & 20.0 & 34.5 & 105.0 & 2.4 & 0.9 & 52.8 & 133 & 123k\\
HumanCLAW$^\dagger$~\cite{siyao2026humanclaw} & Claude Opus 5.5 & 20.0 & 33.3 & 15.0 & 6.7 & 10.0 & 26.8 & 42.1 & 0.8 & 3.6 & 11.9 & 12 & 91k\\
GEMS$^\dagger$~\cite{bao2026generative} & Claude Opus 5.5 & 6.7 & 13.3 & 5.0 & 0.0 & 0.0 & 13.5 & 11.9 & 1.0 & 5.3 & 3.8 & 7 & 98k\\
\midrule
\multirow{5}{*}{Inhabitant} & GLM-5.3-Flash & 44.0 & 80.0 & 10.0 & 0.0 & 70.0 & 53.2 & 36.8 & 1.7 & 5.9 & 38.2 & 112 & 221k\\
 & gpt-6-luna & 30.7 & 40.0 & 25.0 & 6.7 & 50.0 & 41.4 & 45.3 & 1.5 & 5.4 & 46.4 & 122 & 143k\\
 & Gemini 4 Argon & 82.7 & \textbf{100.0} & 85.0 & 46.7 & 80.0 & 84.3 & \textbf{4.0} & \textbf{0.5} & 5.8 & \textbf{12.5} & 73 & 107k\\
 & Claude Sonnet 5.5 & 93.3 & 96.7 & 90.0 & 86.7 & \textbf{100.0} & 94.9 & 17.9 & 1.4 & 6.2 & 23.6 & 69 & 127k\\
 & Claude Opus 5.5 & \textbf{98.7} & \textbf{100.0} & \textbf{100.0} & \textbf{100.0} & 90.0 & \textbf{99.6} & 9.3 & 1.0 & \textbf{4.7} & 17.1 & \textbf{51} & \textbf{99k}\\
\bottomrule
\end{tabular}}
\caption{Results on the 75 tasks of Inhabit. $^\dagger$: baseline adapted to Inhabit. Bold: best success rate; for motion quality and cost, best among the runs of Inhabitant only, since the baselines stop early and return shorter sequences.}
\label{tab:main}
\end{table*}

\subsection{Experimental Setup}
\label{sec:exp_setup}

All experiments run on Inhabit, follow the episode protocol in Sec.~\ref{sec:protocol} and report the metrics defined in Sec.~\ref{sec:metrics}.
The comparisons in Sec.~\ref{sec:main} use all 75 tasks, and the ablations in Secs.~\ref{sec:control_exp} to \ref{sec:loop_exp} use an ablation set of 20 tasks to reduce cost.
It is selected to cover all four tiers (9/4/3/4) and all goal types (Supp.~\ref{sec:supp_tasks}).
In the ablation tables, success and missed grasps are reported as percentages, and the other columns are means per episode.

Unless stated otherwise, Inhabitant uses Kimodo~\cite{rempe2026kimodo} as the motion generator and Claude Opus 5.5 as the language model.
We also run it with Claude Sonnet 5.5, Gemini 4 Argon, gpt-6-luna and the open-weight GLM-5.3-Flash, accessed through their APIs.
A single prompt is used across tasks, language models and settings; each ablation only removes the description of the component that it turns off.

We compare against three agentic HSI baselines: InteractAgent$^\dagger$~\cite{zhang2026interactagent}, which controls a text-to-motion generator with text prompts only; HumanCLAW$^\dagger$~\cite{siyao2026humanclaw}, which calls a fixed set of atomic skills; and GEMS$^\dagger$~\cite{bao2026generative}, which submits one plan and receives no execution feedback.
To evaluate them on Inhabit, we adapt each baseline with minimal changes: each baseline keeps its own control interface and loop, while sharing our language model (Claude Opus 5.5), scene simulator, state description, interaction events, success rule and budget of calls (Supp.~\ref{sec:supp_baselines}).

\subsection{Main Results}
\label{sec:main}

With Claude Opus 5.5, Inhabitant satisfies the success rule on 74 of the 75 tasks, whereas the best baseline using the same language model reaches only 26.7\% success (Tab.~\ref{tab:main}).
The gap widens with task length: the baselines solve some low tier tasks but almost none of the mid- and high-tier tasks.

\begin{figure*}[t]
\centering
\includegraphics[width=\linewidth]{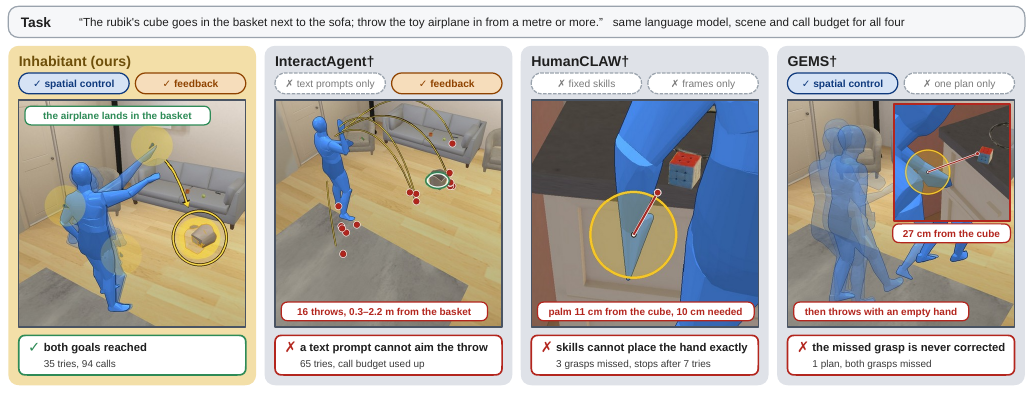}
\caption{Inhabitant and the three baselines on one task with the same language model; chips show which of our two interfaces each agent has and what it has in place of a missing one. Red dots: where the throws of InteractAgent$^\dagger$ land; yellow circle: the 10\,cm radius of the grasp rule; inset: a missed grasp of GEMS$^\dagger$. More examples are in the video on the project page.}
\label{fig:rollouts}
\end{figure*}

Each baseline fails at the interface it lacks.
Since all baselines share the same language model, lacking a specific interface causes each to fail with a distinct cost pattern.
InteractAgent$^\dagger$ has feedback but no spatial control: it can observe what went wrong but can only reword its text prompts. It therefore needs three times as many tries as Inhabitant and most failures use up the budget.
HumanCLAW$^\dagger$ is limited to a fixed set of skills that cannot place a hand precisely enough to interact with an object; the agent recognizes this and stops by itself in every failure, after about a dozen calls.
GEMS$^\dagger$ has spatial control but no feedback: the errors of its single plan accumulate, most of its grasps miss, and it succeeds only on tasks with loose targets such as kicking or throwing.
Fig.~\ref{fig:rollouts} shows these failure modes on one task.

The interfaces matter more than the choice of the language model.
Across five language models, the success of Inhabitant follows model capability: the three stronger models solve at least 82.7\% of the tasks, and the two weaker ones solve less than half, mostly in the low tier and posture tasks.
Nevertheless, both weaker models with Inhabitant's interfaces outperform every baseline with Claude Opus 5.5.
The weaker models gain less from each revision: after a missed grasp, a revised try of Claude Opus 5.5 succeeds twice as often as one of the weaker models (Supp.~\ref{sec:supp_usage}).
As a result, weaker models need more tries, collide more often, and mostly fail by using up the budget.
Inhabitant is also cheaper per solved task: \$1.7 in API fees with Claude Opus 5.5, against \$4.8 to \$21.1 for the baselines (Supp.~\ref{sec:supp_main}).

\subsection{Spatial Control}
\label{sec:control_exp}

\begin{table}[t]
\centering
\small
\setlength{\tabcolsep}{4pt}
\resizebox{\linewidth}{!}{%
\begin{tabular}{lccccc}
\toprule
Control specification & Success & Collisions & Missed grasps & Tries & Calls\\
\midrule
Atomic skills & 30.0 & 8.4 & 68 & 12.6 & 32\\
Text prompts only & 20.0 & 7.2 & 73 & 61.0 & 134\\
+ root waypoints & 90.0 & 9.6 & 62 & 27.9 & 69\\
+ end-effector targets & \textbf{95.0} & \textbf{7.7} & 45 & 17.4 & 51\\
+ full-body keyframes (Inhabitant) & \textbf{95.0} & 7.8 & \textbf{40} & \textbf{14.8} & \textbf{47}\\
\bottomrule
\end{tabular}}
\caption{Granularity of spatial control on the ablation set with Claude Opus 5.5.}
\label{tab:control}
\end{table}

To ablate spatial control, Tab.~\ref{tab:control} starts from text prompts alone and adds the three types of spatial constraints one at a time, with the feedback unchanged.
Without spatial constraints, the agent fails on most tasks, whether it controls motion through text prompts or atomic skills. With text prompts, it spends four times as many tries as the full system.

The three types of constraints help at different levels.
Root waypoints account for most of the success gain, from 20.0\% to 90.0\%: they decide where the body walks and stops, which every task with several steps needs first.
End-effector targets and full-body keyframes then add precision rather than success. They solve only one more task, but missed grasps, the share of the grasp events of all tries that attach no object, fall by a third and the tries per episode by half.
The tasks that fail with root waypoints alone are reaches in which the palm stops short of the object (Supp.~\ref{sec:supp_control}).

Finally, we examine how different language models use the three types of constraints when all of them are available.
In the episodes of Tab.~\ref{tab:main}, we count segments whose control specification contains an end-effector target or a full-body keyframe.
Claude Opus 5.5 uses one in more than a third of its segments, while GLM-5.3-Flash and gpt-6-luna rarely use them and rely mostly on root waypoints.
These two language models also miss grasps more often (Supp.~\ref{sec:supp_usage}).

\subsection{Feedback}
\label{sec:feedback_exp}

\begin{table}[t]
\centering
\small
\setlength{\tabcolsep}{3pt}
\resizebox{\linewidth}{!}{%
\begin{tabular}{cccccccccc}
\toprule
\# & Check & Frames & Body & Scene & Success & Collisions & Missed grasps & Tries & Calls\\
\midrule
1 & \checkmark & \checkmark & & & 85.0 & 14.6 & 58 & 22.7 & 54\\
2 & \checkmark & \checkmark & \checkmark & & \textbf{95.0} & 10.9 & 47 & 19.3 & 64\\
3 & \checkmark & \checkmark & & \checkmark & \textbf{95.0} & 12.4 & 61 & 22.0 & 59\\
4 & \checkmark & \checkmark & \checkmark & \checkmark & \textbf{95.0} & \textbf{7.8} & \textbf{40} & \textbf{14.8} & \textbf{47}\\
5 &  & \checkmark & \checkmark & \checkmark & 85.0 & 9.5 & 47 & 19.8 & 60\\
6 &  & \checkmark & & & 70.0 & 11.1 & 53 & 21.6 & 54\\
\bottomrule
\end{tabular}}
\caption{Granularity of feedback on the ablation set with Claude Opus 5.5 and all three types of spatial constraints. Check, Frames, Body and Scene are the four kinds of feedback of Tab.~\ref{tab:feedback_kinds}.}
\label{tab:feedback}
\end{table}

To ablate feedback, Tab.~\ref{tab:feedback} keeps the full control specification and removes parts of the feedback: first the two kinds of measurements (rows 1 to 4), then the success check (rows 5 and 6).
Without measurement (row 1), the agent solves only two tasks fewer than with all feedback (row 4), but it acts far less precisely: it needs about 50\% more tries, and produces roughly twice as many body collisions.
The rendered frames show that a try failed but not by how much, so the revisions are coarse.

Body and scene measurements complement each other.
Either kind alone (rows 2 and 3) is enough to reach the success rate of the full feedback, but only the two together reduces the tries and the collisions.
Missed grasps fall with the body measurements, which localizes the hand, but not with the scene measurements.

The success check tells the agent after every try whether the task is solved.
Rows 5 and 6 remove this check, requiring the agent to decide when to stop.
With measurements, success drops from 95.0\% to 85.0\% (rows 4 and 5); with rendered frames alone, it drops from 85.0\% to 70.0\% (rows 1 and 6). The typical failure is  premature stopping before the task is actually solved.
On all 75 tasks without the success check, Inhabitant still achieves 86.7\% success, and nine of ten episodes in which it stops by itself are successful (Supp.~\ref{sec:supp_nocheck}).
Thus, quantitative measurements allow the agent to verify the outcome by itself in most episodes.

\subsection{Further Analysis}
\label{sec:loop_exp}

\begin{table}[t]
\centering
\small
\setlength{\tabcolsep}{4pt}
\resizebox{0.7\linewidth}{!}{%
\begin{tabular}{lcccc}
\toprule
Setting & Success & Collisions & Tries & Calls\\
\midrule
Inhabitant & \textbf{95.0} & \textbf{7.8} & 14.8 & 47\\
Single-Try & 75.0 & 30.4 & 13.2 & 36\\
Open-Loop & 0.0 & 10.9 & 4.2 & 2\\
\bottomrule
\end{tabular}}
\caption{The execution loop on the ablation set with Claude Opus 5.5.}
\label{tab:loop}
\end{table}

In Inhabitant, the agent may revise a try using feedback as many times as needed before committing it (Sec.~\ref{sec:loop}).
To assess the value of this revision loop, we compare Inhabitant with two settings that remove it (Tab.~\ref{tab:loop}).
In \emph{Single-Try}, each try is committed as soon as it is executed, so the agent still receives feedback but can only correct a mistake in later segments (similar to ReAct~\cite{yao2022react}).
In \emph{Open-Loop}, the agent writes all control specifications at once, and they are executed without any feedback.
Open-Loop solves no task, since nothing corrects the deviations of the executed motion from the plan.
Single-Try solves four tasks fewer than Inhabitant, and its body collides about four times as often, because every imprecise try stays in the interaction sequence.
Revising a try before committing it therefore helps success and, more importantly, keeps the interaction sequence clean.

The scene simulator moves the body kinematically, without balance or torques.
To test whether the interaction sequences are nevertheless physically plausible, we track those of Inhabitant with Claude Opus 5.5 with the pretrained physics controller PHC+~\cite{luo2023perpetual} on a flat floor, in the parts where the body is not supported by furniture.
The controller follows 94\% of these parts without falling, and half of its failures are in tasks that end in a kneeling or lying posture (Supp.~\ref{sec:supp_phc}).

\section{Conclusion and Future Work}
\label{sec:conclusion}

We presented Inhabitant, a training-free agent framework for long-horizon human-scene interaction that gives the agent what existing agents lack: spatial control over the motion and quantitative feedback on its execution.
To evaluate such interaction, we built Inhabit, a benchmark of 75 household tasks in which every step is carried out by the motion of the whole body and a success rule decides on the physical outcome.
On Inhabit, Inhabitant completes 98.7\% of the tasks with Claude Opus 5.5, where existing agentic HSI baselines complete at most 26.7\% with the same language model, and the ablations confirm that both interfaces are needed: without spatial control the agent fails most tasks, and without quantitative feedback its motion is less precise.

Inhabitant has four main limitations.
First, it requires a scene simulator in which a motion can be tried repeatedly, making it best suited for offline uses such as animation and data generation; an online setting, where every executed motion is irreversible, are better reflected by the Single-Try results in Sec.~\ref{sec:loop_exp}.
Second, the body is driven kinematically, without balance, torques or fingers, and grasping is rule-based; a pretrained physics controller already follows most of the interaction sequences (Sec.~\ref{sec:loop_exp}), and driving the body with such a controller is future work.
Third, the agent is given the state description of the scene and a success check after each try; it still solves most tasks without the check (Sec.~\ref{sec:feedback_exp}), and replacing the given state by perception is left for future work.
Fourth, because of budget constraints, each task is run once and the ablations use a 20-task subset. Moreover, the baselines are our adaptations. Future evaluations should include repeated runs and comparisons with the original implementations. 

\section*{LLM Usage Statement}

LLMs are an object of study in this work: the agent of Inhabitant is an LLM that is used without training. We evaluate Claude Opus 5.5, Claude Sonnet 5.5, Gemini 4 Argon, gpt-6-luna and the open-weight GLM-5.3-Flash as this agent, and Claude Opus 5.5 as the agent of the adapted baselines (Sec.~\ref{sec:exp_setup}). In addition, we used Claude as an assistant for writing and debugging the code of the agent, the benchmark and the evaluation, for launching and monitoring experiment runs and computing the reported statistics, for surveying related work, and for drafting and editing parts of the manuscript text and figures. The research questions, method design, benchmark tasks, experimental protocol and conclusions are the authors' own. All LLM-assisted code, text, figures and citations were reviewed and verified by the authors, and all reported results come from our own runs.

{
    \small
    \bibliographystyle{ieeenat_fullname}
    \bibliography{main}

@article{starke2019neural,
  title={Neural state machine for character-scene interactions},
  author={Starke, Sebastian and Zhang, He and Komura, Taku and Saito, Jun},
  journal={ACM Transactions on Graphics},
  volume={38},
  number={6},
  pages={178},
  year={2019},
  publisher={Association for Computing Machinery (ACM)}
}

@inproceedings{hassan2021stochastic,
  title={Stochastic scene-aware motion prediction},
  author={Hassan, Mohamed and Ceylan, Duygu and Villegas, Ruben and Saito, Jun and Yang, Jimei and Zhou, Yi and Black, Michael J},
  booktitle={Proceedings of the IEEE/CVF International Conference on Computer Vision},
  pages={11374--11384},
  year={2021}
}

@inproceedings{hassan2023synthesizing,
  title={Synthesizing physical character-scene interactions},
  author={Hassan, Mohamed and Guo, Yunrong and Wang, Tingwu and Black, Michael and Fidler, Sanja and Peng, Xue Bin},
  booktitle={ACM SIGGRAPH 2023 Conference Proceedings},
  pages={1--9},
  year={2023}
}

@inproceedings{puig2024habitat,
  title={Habitat 3.0: A co-habitat for humans, avatars, and robots},
  author={Puig, Xavier and Undersander, Eric and Szot, Andrew and Dallaire Cote, Mikael and Yang, Tsung-Yen and Partsey, Ruslan and Desai, Ruta and Clegg, Alexander and Hlavac, Michal and Min, So Yeon and others},
  booktitle={International Conference on Learning Representations},
  volume={2024},
  pages={15306--15336},
  year={2024}
}

@inproceedings{zhang2022couch,
  title={Couch: Towards controllable human-chair interactions},
  author={Zhang, Xiaohan and Bhatnagar, Bharat Lal and Starke, Sebastian and Guzov, Vladimir and Pons-Moll, Gerard},
  booktitle={European Conference on Computer Vision},
  pages={518--535},
  year={2022},
  organization={Springer}
}

@article{wang2022humanise,
  title={Humanise: Language-conditioned human motion generation in 3d scenes},
  author={Wang, Zan and Chen, Yixin and Liu, Tengyu and Zhu, Yixin and Liang, Wei and Huang, Siyuan},
  journal={Advances in Neural Information Processing Systems},
  volume={35},
  pages={14959--14971},
  year={2022}
}

@inproceedings{zhao2023synthesizing,
  title={Synthesizing diverse human motions in 3d indoor scenes},
  author={Zhao, Kaifeng and Zhang, Yan and Wang, Shaofei and Beeler, Thabo and Tang, Siyu},
  booktitle={2023 IEEE/CVF International Conference on Computer Vision (ICCV)},
  pages={14692--14703},
  year={2023},
  organization={IEEE}
}

@inproceedings{li2024controllable,
  title={Controllable human-object interaction synthesis},
  author={Li, Jiaman and Clegg, Alexander and Mottaghi, Roozbeh and Wu, Jiajun and Puig, Xavier and Liu, C Karen},
  booktitle={European Conference on Computer Vision},
  pages={54--72},
  year={2024},
  organization={Springer}
}

@inproceedings{jiang2024scaling,
  title={Scaling up dynamic human-scene interaction modeling},
  author={Jiang, Nan and Zhang, Zhiyuan and Li, Hongjie and Ma, Xiaoxuan and Wang, Zan and Chen, Yixin and Liu, Tengyu and Zhu, Yixin and Huang, Siyuan},
  booktitle={2024 IEEE/CVF Conference on Computer Vision and Pattern Recognition (CVPR)},
  pages={1737--1747},
  year={2024},
  organization={IEEE}
}

@inproceedings{pan2025tokenhsi,
  title={Tokenhsi: Unified synthesis of physical human-scene interactions through task tokenization},
  author={Pan, Liang and Yang, Zeshi and Dou, Zhiyang and Wang, Wenjia and Huang, Buzhen and Dai, Bo and Komura, Taku and Wang, Jingbo},
  booktitle={2025 IEEE/CVF Conference on Computer Vision and Pattern Recognition (CVPR)},
  pages={5379--5391},
  year={2025},
  organization={IEEE}
}

@inproceedings{xiao2024unified,
  title={Unified human-scene interaction via prompted chain-of-contacts},
  author={Xiao, Zeqi and Wang, Tai and Wang, Jingbo and Cao, Jinkun and Zhang, Wenwei and Dai, Bo and Lin, Dahua and Pang, Jiangmiao},
  booktitle={International Conference on Learning Representations},
  volume={2024},
  pages={24450--24461},
  year={2024}
}

@inproceedings{wu2025human,
  title={Human-object interaction from human-level instructions},
  author={Wu, Zhen and Li, Jiaman and Xu, Pei and Liu, C Karen},
  booktitle={2025 IEEE/CVF International Conference on Computer Vision (ICCV)},
  pages={11176--11186},
  year={2025},
  organization={IEEE}
}

@inproceedings{wang2025sims,
  title={Sims: Simulating stylized human-scene interactions with retrieval-augmented script generation},
  author={Wang, Wenjia and Pan, Liang and Dou, Zhiyang and Mei, Jidong and Liao, Zhouyingcheng and Lou, Yuke and Wu, Yifan and Yang, Lei and Wang, Jingbo and Komura, Taku},
  booktitle={2025 IEEE/CVF International Conference on Computer Vision (ICCV)},
  pages={14117--14127},
  year={2025},
  organization={IEEE}
}

@inproceedings{jian2026endowing,
  title={Endowing GPT-4 with a humanoid body: Building the bridge between off-the-shelf VLMs and the physical world},
  author={Jian, Yingzhao and Wang, Zhongan and Yang, Yi and Fan, Hehe},
  booktitle={International Conference on Learning Representations},
  volume={2026},
  pages={134007--134065},
  year={2026}
}

@article{bao2026generative,
  title={Generative Embodied Multiple Behavior Control Systems for Human-like Agents},
  author={Bao, Chongyu and Yang, Haokai and Wang, Yuhan and An, Zhaochong and Liu, Kunpeng and Liu, Xiaolan},
  journal={arXiv preprint arXiv:2609.22691},
  year={2026}
}

@article{zhang2026interactagent,
  title={InteractAgent: Agentic Human Motion Interaction with Memory-Augmented Reflection},
  author={Zhang, Tao and Zhang, Senhe and Zhang, Zeyu and Zhang, Yujia and Gong, Dong},
  journal={arXiv preprint arXiv:2610.25649},
  year={2026}
}

@article{siyao2026humanclaw,
  title={HumanCLAW: Can Vision-Language Models Act Through a Body?},
  author={Siyao, Li and Gu, Jiawei and Liu, Shuai and Hu, Kairui and Li, Zekun and Li, Linjie and Tang, Chengcheng and Wu, Po-Chen and Shugurov, Ivan and Ma, Lingni and others},
  journal={arXiv preprint arXiv:2607.27180},
  year={2026}
}

@inproceedings{wu2025motion,
  title={Motion-agent: A conversational framework for human motion generation with llms},
  author={Wu, Qi and Zhao, Yubo and Wang, Yifan and Liu, Xinhang and Tai, Yu-Wing and Tang, Chi-Keung},
  booktitle={International Conference on Learning Representations},
  volume={2025},
  pages={48023--48043},
  year={2025}
}

@article{qu2024gpt,
  title={Gpt-connect: Interaction between text-driven human motion generator and 3d scenes in a training-free manner},
  author={Qu, Haoxuan and Guo, Ziyan and Liu, Jun},
  journal={arXiv preprint arXiv:2403.14947},
  year={2024}
}

@article{rempe2026kimodo,
  title={Kimodo: Scaling controllable human motion generation},
  author={Rempe, Davis and Petrovich, Mathis and Yuan, Ye and Zhang, Haotian and Peng, Xue Bin and Jiang, Yifeng and Wang, Tingwu and Iqbal, Umar and Minor, David and de Ruyter, Michael and others},
  journal={arXiv preprint arXiv:2603.15546},
  year={2026}
}

@article{siyao2025half,
  title={Half-physics: Enabling kinematic 3d human model with physical interactions},
  author={Siyao, Li and Feng, Yao and Taheri, Omid and Loy, Chen Change and Black, Michael J},
  journal={arXiv preprint arXiv:2507.23778},
  year={2025}
}

@article{szot2021habitat,
  title={Habitat 2.0: Training home assistants to rearrange their habitat},
  author={Szot, Andrew and Clegg, Alexander and Undersander, Eric and Wijmans, Erik and Zhao, Yili and Turner, John and Maestre, Noah and Mukadam, Mustafa and Chaplot, Devendra Singh and Maksymets, Oleksandr and others},
  journal={Advances in neural information processing systems},
  volume={34},
  pages={251--266},
  year={2021}
}

@article{yang2025embodiedbench,
  title={Embodiedbench: Comprehensive benchmarking multi-modal large language models for vision-driven embodied agents},
  author={Yang, Rui and Chen, Hanyang and Zhang, Junyu and Zhao, Mark and Qian, Cheng and Wang, Kangrui and Wang, Qineng and Koripella, Teja Venkat and Movahedi, Marziyeh and Li, Manling and others},
  journal={arXiv preprint arXiv:2502.09560},
  year={2025}
}

@inproceedings{chang2025partnr,
  title={Partnr: A benchmark for planning and reasoning in embodied multi-agent tasks},
  author={Chang, Matthew and Chhablani, Gunjan and Clegg, Alexander and Dallaire Cote, Mikael and Desai, Ruta and Hlavac, Michal and Karashchuk, Vladimir and Krantz, Jacob and Mottaghi, Roozbeh and Parashar, Priyam and others},
  booktitle={International Conference on Learning Representations},
  volume={2025},
  pages={65205--65268},
  year={2025}
}

@inproceedings{zhang2022wanderings,
  title={The wanderings of odysseus in 3d scenes},
  author={Zhang, Yan and Tang, Siyu},
  booktitle={2022 IEEE/CVF Conference on Computer Vision and Pattern Recognition (CVPR)},
  pages={20449--20459},
  year={2022},
  organization={IEEE}
}

@inproceedings{yi2024generating,
  title={Generating human interaction motions in scenes with text control},
  author={Yi, Hongwei and Thies, Justus and Black, Michael J and Peng, Xue Bin and Rempe, Davis},
  booktitle={European Conference on Computer Vision},
  pages={246--263},
  year={2024},
  organization={Springer}
}

@inproceedings{wang2024move,
  title={Move as you say, interact as you can: Language-guided human motion generation with scene affordance},
  author={Wang, Zan and Chen, Yixin and Jia, Baoxiong and Li, Puhao and Zhang, Jinlu and Zhang, Jingze and Liu, Tengyu and Zhu, Yixin and Liang, Wei and Huang, Siyuan},
  booktitle={2024 IEEE/CVF Conference on Computer Vision and Pattern Recognition (CVPR)},
  pages={433--444},
  year={2024},
  organization={IEEE}
}

@inproceedings{jiang2024autonomous,
  title={Autonomous character-scene interaction synthesis from text instruction},
  author={Jiang, Nan and He, Zimo and Wang, Zi and Li, Hongjie and Chen, Yixin and Huang, Siyuan and Zhu, Yixin},
  booktitle={SIGGRAPH Asia 2024 Conference Papers},
  pages={1--11},
  year={2024}
}

@inproceedings{wang2025hsi,
  title={Hsi-gpt: A general-purpose large scene-motion-language model for human scene interaction},
  author={Wang, Yuan and Li, Yali and Li, Xiang and Wang, Shengjin},
  booktitle={2025 IEEE/CVF Conference on Computer Vision and Pattern Recognition (CVPR)},
  pages={7147--7157},
  year={2025},
  organization={IEEE}
}

@article{wang2023physhoi,
  title={Physhoi: Physics-based imitation of dynamic human-object interaction},
  author={Wang, Yinhuai and Lin, Jing and Zeng, Ailing and Luo, Zhengyi and Zhang, Jian and Zhang, Lei},
  journal={arXiv preprint arXiv:2312.04393},
  year={2023}
}

@inproceedings{wang2025skillmimic,
  title={Skillmimic: Learning basketball interaction skills from demonstrations},
  author={Wang, Yinhuai and Zhao, Qihan and Yu, Runyi and Tsui, Hok Wai and Zeng, Ailing and Lin, Jing and Luo, Zhengyi and Yu, Jiwen and Li, Xiu and Chen, Qifeng and others},
  booktitle={2025 IEEE/CVF Conference on Computer Vision and Pattern Recognition (CVPR)},
  pages={17540--17549},
  year={2025},
  organization={IEEE}
}

@inproceedings{xu2025intermimic,
  title={Intermimic: Towards universal whole-body control for physics-based human-object interactions},
  author={Xu, Sirui and Ling, Hung Yu and Wang, Yu-Xiong and Gui, Liang-Yan},
  booktitle={2025 IEEE/CVF Conference on Computer Vision and Pattern Recognition (CVPR)},
  pages={12266--12277},
  year={2025},
  organization={IEEE}
}

@article{tessler2024maskedmimic,
  title={Maskedmimic: Unified physics-based character control through masked motion inpainting},
  author={Tessler, Chen and Guo, Yunrong and Nabati, Ofir and Chechik, Gal and Peng, Xue Bin},
  journal={ACM Transactions On Graphics (TOG)},
  volume={43},
  number={6},
  pages={1--21},
  year={2024},
  publisher={ACM, New York, NY, USA}
}

@inproceedings{wu2025uniphys,
  title={Uniphys: Unified planner and controller with diffusion for flexible physics-based character control},
  author={Wu, Yan and Karunratanakul, Korrawe and Luo, Zhengyi and Tang, Siyu},
  booktitle={2025 IEEE/CVF International Conference on Computer Vision (ICCV)},
  pages={13214--13224},
  year={2025},
  organization={IEEE}
}

@article{xu2024humanvla,
  title={Humanvla: Towards vision-language directed object rearrangement by physical humanoid},
  author={Xu, Xinyu and Zhang, Yizheng and Li, Yong-Lu and Han, Lei and Lu, Cewu},
  journal={Advances in Neural Information Processing Systems},
  volume={37},
  pages={18633--18659},
  year={2024}
}

@article{zhang2025lhm,
  title={LHM-Humanoid: Long-Horizon Human Motion Control for Continuous Object Transport in Cluttered Scenes},
  author={Zhang, Haozhuo and Sun, Jingkai and Caprio, Michele and Cangelosi, Angelo and Tang, Jian and Zhang, Shanghang and Zhang, Qiang and Pan, Wei},
  journal={arXiv preprint arXiv:2508.16943},
  year={2025}
}

@inproceedings{xie2024omnicontrol,
  title={Omnicontrol: Control any joint at any time for human motion generation},
  author={Xie, Yiming and Jampani, Varun and Zhong, Lei and Sun, Deqing and Jiang, Huaizu},
  booktitle={International Conference on Learning Representations},
  volume={2024},
  pages={28176--28194},
  year={2024}
}

@inproceedings{karunratanakul2023guided,
  title={Guided motion diffusion for controllable human motion synthesis},
  author={Karunratanakul, Korrawe and Preechakul, Konpat and Suwajanakorn, Supasorn and Tang, Siyu},
  booktitle={2023 IEEE/CVF International Conference on Computer Vision (ICCV)},
  pages={2151--2162},
  year={2023},
  organization={IEEE}
}

@inproceedings{wan2024tlcontrol,
  title={Tlcontrol: Trajectory and language control for human motion synthesis},
  author={Wan, Weilin and Dou, Zhiyang and Komura, Taku and Wang, Wenping and Jayaraman, Dinesh and Liu, Lingjie},
  booktitle={European Conference on Computer Vision},
  pages={37--54},
  year={2024},
  organization={Springer}
}

@inproceedings{cohan2024flexible,
  title={Flexible motion in-betweening with diffusion models},
  author={Cohan, Setareh and Tevet, Guy and Reda, Daniele and Peng, Xue Bin and Van de Panne, Michiel},
  booktitle={ACM SIGGRAPH 2024 conference papers},
  pages={1--9},
  year={2024}
}

@inproceedings{karunratanakul2024optimizing,
  title={Optimizing diffusion noise can serve as universal motion priors},
  author={Karunratanakul, Korrawe and Preechakul, Konpat and Aksan, Emre and Beeler, Thabo and Suwajanakorn, Supasorn and Tang, Siyu},
  booktitle={2024 IEEE/CVF Conference on Computer Vision and Pattern Recognition (CVPR)},
  pages={1334--1345},
  year={2024},
  organization={IEEE}
}

@inproceedings{pinyoanuntapong2025maskcontrol,
  title={Maskcontrol: Spatio-temporal control for masked motion synthesis},
  author={Pinyoanuntapong, Ekkasit and Saleem, Muhammad and Karunratanakul, Korrawe and Wang, Pu and Xue, Hongfei and Chen, Chen and Guo, Chuan and Cao, Junli and Ren, Jian and Tulyakov, Sergey},
  booktitle={Proceedings of the IEEE/CVF International Conference on Computer Vision},
  pages={9955--9965},
  year={2025}
}

@inproceedings{zhao2025dartcontrol,
  title={DartControl: A diffusion-based autoregressive motion model for real-time text-driven motion control},
  author={Zhao, Kaifeng and Li, Gen and Tang, Siyu},
  booktitle={International Conference on Learning Representations},
  volume={2025},
  pages={23569--23592},
  year={2025}
}

@inproceedings{li2025genmo,
  title={Genmo: A generalist model for human motion},
  author={Li, Jiefeng and Cao, Jinkun and Zhang, Haotian and Rempe, Davis and Kautz, Jan and Iqbal, Umar and Yuan, Ye},
  booktitle={2025 IEEE/CVF International Conference on Computer Vision (ICCV)},
  pages={11766--11776},
  year={2025},
  organization={IEEE}
}

@inproceedings{petrovich2024multi,
  title={Multi-track timeline control for text-driven 3d human motion generation},
  author={Petrovich, Mathis and Litany, Or and Iqbal, Umar and Black, Michael J and Varol, Gul and Bin Peng, Xue and Rempe, Davis},
  booktitle={Proceedings of the IEEE/CVF Conference on Computer Vision and Pattern Recognition},
  pages={1911--1921},
  year={2024}
}

@inproceedings{barquero2024seamless,
  title={Seamless human motion composition with blended positional encodings},
  author={Barquero, German and Escalera, Sergio and Palmero, Cristina},
  booktitle={2024 IEEE/CVF Conference on Computer Vision and Pattern Recognition (CVPR)},
  pages={457--469},
  year={2024},
  organization={IEEE}
}

@inproceedings{goel2024iterative,
  title={Iterative motion editing with natural language},
  author={Goel, Purvi and Wang, Kuan-Chieh and Liu, C Karen and Fatahalian, Kayvon},
  booktitle={ACM SIGGRAPH 2024 Conference Papers},
  pages={1--9},
  year={2024}
}

@inproceedings{liu2024programmable,
  title={Programmable motion generation for open-set motion control tasks},
  author={Liu, Hanchao and Zhan, Xiaohang and Huang, Shaoli and Mu, Tai-Jiang and Shan, Ying},
  booktitle={2024 IEEE/CVF Conference on Computer Vision and Pattern Recognition (CVPR)},
  pages={1399--1408},
  year={2024},
  organization={IEEE}
}

@article{zheng2026re,
  title={{Re$^2$MoGen}: Open-Vocabulary Motion Generation via LLM Reasoning and Physics-Aware Refinement},
  author={Zheng, Jiakun and Xiao, Ting and Cao, Shiqin and Li, Xinran and Wang, Zhe and Bai, Chenjia},
  journal={arXiv preprint arXiv:2604.17807},
  year={2026}
}

@article{li2026world2hosi,
  title={World2HOSI: A Multi-Expert Harness for Long-Horizon Human-Object-Scene Interaction Generation},
  author={Li, Jiushu and Wang, Haoran and Liu, Yumeng and Dou, Zhiyang and Liu, Yuan and Ma, Yuexin and Lin, Cheng},
  journal={arXiv preprint arXiv:2610.27725},
  year={2026}
}

@inproceedings{zhang2026interpose,
  title={InterPose: Learning to Generate Human-Object Interactions from Large-Scale Web Videos},
  author={Zhang, Yangsong and Butt, Abdul Ahad and Varol, G{\"u}l and Laptev, Ivan},
  booktitle={2026 International Conference on 3D Vision (3DV)},
  pages={750--759},
  year={2026},
  organization={IEEE}
}

@inproceedings{puig2018virtualhome,
  title={Virtualhome: Simulating household activities via programs},
  author={Puig, Xavier and Ra, Kevin and Boben, Marko and Li, Jiaman and Wang, Tingwu and Fidler, Sanja and Torralba, Antonio},
  booktitle={2018 IEEE/cvf conference on computer vision and pattern recognition},
  pages={8494--8502},
  year={2018},
  organization={IEEE}
}

@inproceedings{savva2019habitat,
  title={Habitat: A platform for embodied ai research},
  author={Savva, Manolis and Kadian, Abhishek and Maksymets, Oleksandr and Zhao, Yili and Wijmans, Erik and Jain, Bhavana and Straub, Julian and Liu, Jia and Koltun, Vladlen and Malik, Jitendra and others},
  booktitle={2019 IEEE/CVF International Conference on Computer Vision (ICCV)},
  pages={9338--9346},
  year={2019},
  organization={IEEE}
}

@inproceedings{shridhar2020alfred,
  title={Alfred: A benchmark for interpreting grounded instructions for everyday tasks},
  author={Shridhar, Mohit and Thomason, Jesse and Gordon, Daniel and Bisk, Yonatan and Han, Winson and Mottaghi, Roozbeh and Zettlemoyer, Luke and Fox, Dieter},
  booktitle={2020 IEEE/CVF conference on computer vision and pattern recognition (CVPR)},
  pages={10737--10746},
  year={2020},
  organization={IEEE}
}

@inproceedings{li2023behavior,
  title={Behavior-1k: A benchmark for embodied ai with 1,000 everyday activities and realistic simulation},
  author={Li, Chengshu and Zhang, Ruohan and Wong, Josiah and Gokmen, Cem and Srivastava, Sanjana and Mart{\'\i}n-Mart{\'\i}n, Roberto and Wang, Chen and Levine, Gabrael and Lingelbach, Michael and Sun, Jiankai and others},
  booktitle={Conference on Robot Learning},
  pages={80--93},
  year={2023},
  organization={PMLR}
}

@article{li2024embodied,
  title={Embodied agent interface: Benchmarking llms for embodied decision making},
  author={Li, Manling and Zhao, Shiyu and Wang, Qineng and Wang, Kangrui and Zhou, Yu and Srivastava, Sanjana and Gokmen, Cem and Lee, Tony and Li, Li E and Zhang, Ruohan and others},
  journal={Advances in Neural Information Processing Systems},
  volume={37},
  pages={100428--100534},
  year={2024}
}

@inproceedings{shukla2025maniskill,
  title={Maniskill-hab: A benchmark for low-level manipulation in home rearrangement tasks},
  author={Shukla, Arth and Tao, Stone and Su, Hao},
  booktitle={International Conference on Learning Representations},
  volume={2025},
  pages={15288--15317},
  year={2025}
}

@article{yao2022react,
  title={React: Synergizing reasoning and acting in language models},
  author={Yao, Shunyu and Zhao, Jeffrey and Yu, Dian and Du, Nan and Shafran, Izhak and Narasimhan, Karthik and Cao, Yuan},
  journal={arXiv preprint arXiv:2210.03629},
  year={2022}
}

@article{huang2022inner,
  title={Inner monologue: Embodied reasoning through planning with language models},
  author={Huang, Wenlong and Xia, Fei and Xiao, Ted and Chan, Harris and Liang, Jacky and Florence, Pete and Zeng, Andy and Tompson, Jonathan and Mordatch, Igor and Chebotar, Yevgen and others},
  journal={arXiv preprint arXiv:2207.05608},
  year={2022}
}

@inproceedings{liang2023code,
  title={Code as policies: Language model programs for embodied control},
  author={Liang, Jacky and Huang, Wenlong and Xia, Fei and Xu, Peng and Hausman, Karol and Ichter, Brian and Florence, Pete and Zeng, Andy},
  booktitle={2023 IEEE International conference on robotics and automation (ICRA)},
  pages={9493--9500},
  year={2023},
  organization={IEEE}
}

@article{huang2023voxposer,
  title={Voxposer: Composable 3d value maps for robotic manipulation with language models},
  author={Huang, Wenlong and Wang, Chen and Zhang, Ruohan and Li, Yunzhu and Wu, Jiajun and Fei-Fei, Li},
  journal={arXiv preprint arXiv:2307.05973},
  year={2023}
}

@article{huang2024rekep,
  title={Rekep: Spatio-temporal reasoning of relational keypoint constraints for robotic manipulation},
  author={Huang, Wenlong and Wang, Chen and Li, Yunzhu and Zhang, Ruohan and Fei-Fei, Li},
  journal={arXiv preprint arXiv:2409.01652},
  year={2024}
}

@article{rana2023sayplan,
 title={Sayplan: Grounding large language models using 3d scene graphs for scalable robot task planning},
 author={Rana, Krishan and Haviland, Jesse and Garg, Sourav and Abou-Chakra, Jad and Reid, Ian and Suenderhauf, Niko},
 journal={arXiv preprint arXiv:2307.06135},
 year={2023}
}

@article{tsui2026demonstration,
 title={Demonstration-free robotic control via LLM agents},
 author={Tsui, Brian Y and Fang, Alan Y and Hwu, Tiffany J},
 journal={arXiv preprint arXiv:2601.20334},
 year={2026}
}

@inproceedings{luo2023perpetual,
 title={Perpetual humanoid control for real-time simulated avatars},
 author={Luo, Zhengyi and Cao, Jinkun and Kitani, Kris and Xu, Weipeng and others},
 booktitle={Proceedings of the IEEE/CVF International Conference on Computer Vision},
 pages={10895--10904},
 year={2023}
}

@inproceedings{deng2026human,
  title={Human-object interaction via automatically designed vlm-guided motion policy},
  author={Deng, Zekai and Shi, Ye and Ji, Kaiyang and Xu, Lan and Huang, Shaoli and Wang, Jingya},
  booktitle={International Conference on Learning Representations},
  volume={2026},
  pages={100691--100718},
  year={2026}
}
}

\clearpage
\appendix
\maketitlesupplementary



\section{Details of Inhabitant}
\label{sec:supp_method}
This section supplements Sec.~\ref{sec:method}.

\subsection{The loop of the agent}
\label{sec:supp_loop}
Algorithm~\ref{alg:loop} gives the loop of Sec.~\ref{sec:loop} as pseudocode.

\begin{algorithm}[h]
\caption{Building an interaction sequence with Inhabitant}
\label{alg:loop}
\small
\begin{algorithmic}[1]
\Require instruction $I$, initial state $s_0$, budget $B$
\Ensure interaction sequence $\Sigma = (\tau_1, \dots, \tau_K)$
\State $s \gets s_0$, \ $\Sigma \gets (\,)$, \ $\tau \gets \varnothing$, \ $H \gets \varnothing$
\For{$n = 1, \dots, B$} \Comment{one call per step}
  \State $a \gets \pi(I, \phi(s), H)$
  \If{$a = \textsc{try}(c, e)$} \Comment{spatial control}
    \State $\tau \gets \mathcal{E}\big(s, \rho(\mathcal{G}(c; h)), e\big)$ \Comment{$h$: end of $\Sigma$}
    \State $r \gets o(\tau)$ \Comment{qualitative feedback}
  \ElsIf{$a = \textsc{measure}(q)$}
    \State $r \gets M(q, \tau)$ \Comment{quantitative feedback}
  \ElsIf{$a = \textsc{commit}$ and $\tau \neq \varnothing$}
    \State append $\tau$ to $\Sigma$, \ $s \gets$ last state of $\tau$, \ $\tau \gets \varnothing$
    \If{$G(\Sigma) = 1$} \Return $\Sigma$ \EndIf
    \State $r \gets \phi(s)$
  \ElsIf{$a = \textsc{rollback}(j)$}
    \State keep $\tau_1, \dots, \tau_j$ in $\Sigma$, \ $s \gets s_{t_j}$, \ $\tau \gets \varnothing$
    \State $r \gets \phi(s)$
  \Else \ \Return $\Sigma$ \Comment{the agent stops}
  \EndIf
  \State $H \gets H \cup \{(a, r)\}$
\EndFor
\State \Return $\Sigma$
\end{algorithmic}
\end{algorithm}

\subsection{Prompt and tool definitions}
\label{sec:supp_prompts}
The prompt of the agent (Sec.~\ref{sec:loop}) consists of a system prompt and a first message with the instruction and the state description $\phi(s_0)$ of the initial state.
The system prompt has two parts.
The first part, written by us, is given verbatim in Figs.~\ref{fig:supp_prompt_a} and \ref{fig:supp_prompt_b}: it describes the inputs of the motion generator, the task of the agent, the tools, the map, the rule of the interaction events and the dimensions of the simulated body.
The second part is the public documentation of Kimodo, copied unchanged: its pages on the constraints, on their JSON format (coordinate space and units, constraint types, examples), on the generation parameters and on best practices (text prompting, constraints, limitations), about 11{,}000 characters.
The system prompt is the same for every task and every language model, and it names no object, no piece of furniture and no task.

\begin{figure*}[tp]
\begin{promptbox}{System prompt (part 1 of 2)}
# Moving with Kimodo

You move by asking a motion generator, Kimodo, for one segment of motion at a time. Its own documentation is at the end ("Kimodo's documentation"). generate_segment takes Kimodo's inputs:
- text: a description of the motion, or a list of descriptions that are generated one after another;
- seconds: its length (one per text); a segment is 0.5 to 8 s, each text at most 10 s;
- constraints: Kimodo's constraint list (its JSON format), in Kimodo's frame of this segment;
- seed (default 0; the same inputs with the same seed give the same motion; its execution in the simulator can still differ by a few cm between identical tries), post (post-processing, default true), cfg_weight ([text, constraint], default [2, 2]).
Kimodo's frame of a segment: the origin is on the floor below your pelvis at the start of the segment, +Z is the direction you face at the start, Y is up, X is to your left; frame_indices count frames from the start of the segment, 30 per second. The other tools report positions as forward / left of your pelvis and heights above the floor: Kimodo's Z / X / Y. The skeleton has 22 joints, in this order (one rotation each in local_joints_rot): pelvis, left_hip, right_hip, spine1, left_knee, right_knee, spine2, left_ankle, right_ankle, spine3, left_foot, right_foot, neck, left_collar, right_collar, head, left_shoulder, right_shoulder, left_elbow, right_elbow, left_wrist, right_wrist. local_joints_rot are SMPL-X's local joint rotations: with all of them zero the body stands in a T-pose facing +Z, arms along X. global_root_heading [cos theta, sin theta] is the facing as its (Z, X) components: theta = 0 faces +Z, theta = 90 deg faces +X (turned to the left). A hand constraint fixes the wrist (position and rotation), a foot constraint the ankle and the toe joint; both also fix the root's position, height and facing at that frame. Every segment starts from your actual pose: Kimodo gets the last frames of what your body did as a full-body keyframe just before the segment. The generated motion is fitted to your body, placed in the scene and executed in a physics simulator.

## Your task

You are the mind of a person in an apartment, and you must complete a household task by moving your own body. Nothing happens by itself: every movement is a segment of motion that you have Kimodo make (above). You act only through the tools. Move as a real person would: do not bump into furniture, walls, plants or other things (only your hands touch what you grasp or hold), and do not let your feet slide on the floor.

### Tools
- generate_segment(intent, text, seconds, constraints, seed, post, cfg_weight, events): Kimodo makes one segment from the last committed state, executed in a physics simulator (furniture and walls stop the body). It returns whether the goal is now satisfied and a video of the try (left: a camera following you; right: a side camera looking at you and at the target of the segment); measure anything else with inspect. Nothing is kept until you commit: every try starts again from the last committed state.
- commit(): keeps the last try; the world continues from where it ended. Returns the new state near you (within 2 m; get_state() gives all of it).
- get_state(): the current state.
- retry_from(i): undo: puts the world back to the start of committed segment i and drops segments i..end; then try / commit as before.
- inspect(queries): read-only measurements of the world and of your tries: where things and body parts are (now or over a try), distances over time, body contacts, the size of any thing, an executed pose in Kimodo's format and the joint positions of a pose in Kimodo's format. It changes nothing.
The episode ends when a segment you commit completes the task.

After each try, look at the result and the video and change the inputs until the segment does what you intended; then commit it.
\end{promptbox}
\caption{System prompt of Inhabitant, part 1 of 2: the inputs of the motion generator, the task and the tools. The text is verbatim, with the symbols for the angle and the degree spelled out. It is followed by Fig.~\ref{fig:supp_prompt_b} and by the documentation of Kimodo.}
\label{fig:supp_prompt_a}
\end{figure*}

\begin{figure*}[tp]
\begin{promptbox}{System prompt (part 2 of 2)}
### The room
- In the top-down map, '.' means a body 0.2 m in radius around the pelvis fits there; your shoulders and swinging arms reach further.
- The state lists all places, objects and handles with their distance and direction from you.

### Rules of this world
- Pick up: a grasp event names a hand, an object id and a time window. If, inside the window, the centre of that palm comes within 0.10 m of the object (its bounding box) for 0.1 s, the object attaches to that hand and moves with it.
- Put down: with the object in hand, a release event names the hand, where it should go (a surface id, an object id such as a basket or box, or "floor") and the time you let go. The object then leaves the hand with the velocity it had there (so it can be thrown), falls and rests where physics puts it; it counts as on a surface if it ends inside that surface's region, and in a basket or box if it ends inside it.
- Drawers and the fridge: grasp the handle id (palm within 0.12 m of the handle), then move that hand along the given opening direction; the drawer / door follows the hand while it is held. Release it afterwards (name any surface; not checked).
- After the motion of a segment the world runs on for 0.3 s (1.5 s after a release) with your body held still, so that things settle; the times of a try include it.
- One hand holds one thing. Whether the task is accomplished is checked after every try.

### Your body
Your body (the generated motion is fitted to it; the generator's own body is a little taller): standing heights above the floor: pelvis 0.98 m, knees 0.51 m, shoulders 1.41 m, head 1.60 m; the palm centre, which the grasp rule uses, is 0.054 m beyond the wrist.
\end{promptbox}
\caption{System prompt of Inhabitant, part 2 of 2: the map, the rule of the interaction events and the simulated body.}
\label{fig:supp_prompt_b}
\end{figure*}

Tab.~\ref{tab:supp_tools} lists the tools that implement the four calls of Eq.~\eqref{eq:control}, and Tab.~\ref{tab:supp_queries} the queries of a measurement.
A measurement may contain several queries, and a query refers to the last try or, with an index, to a committed segment.
Besides the four calls, the agent can request the state description of the current state at any time.
Fig.~\ref{fig:supp_example} shows a try, a measurement and a commit of one episode as the agent issues them and as they are answered.

\begin{table}[h]
\centering
\small
\setlength{\tabcolsep}{3pt}
\begin{tabular}{@{}lp{0.36\linewidth}p{0.36\linewidth}@{}}
\toprule
Call & Tool and arguments & Result\\
\midrule
$\textsc{try}(c, e)$ & generate\_segment: text prompts, durations, constraints, seed, post-processing, guidance weights ($c$) and interaction events ($e$) & success check and eight rendered frames of the try\\
$\textsc{measure}(q)$ & inspect: a list of queries & one value or time series per query\\
$\textsc{commit}$ & commit & state description within 2\,m of the body\\
$\textsc{rollback}(j)$ & retry\_from: index of a committed segment & state description\\
-- & get\_state & full state description\\
\bottomrule
\end{tabular}
\caption{The tools of Inhabitant and the calls of Eq.~\eqref{eq:control} that they implement.}
\label{tab:supp_tools}
\end{table}

\begin{table}[h]
\centering
\small
\setlength{\tabcolsep}{3pt}
\begin{tabular}{@{}llp{0.6\linewidth}@{}}
\toprule
Query & Kind & Returns\\
\midrule
where & body, scene & position of a body part or an object, now, at a time of the try or every 0.1\,s; for an object also the surface or hand it is on, for a handle whether it is open\\
distance & body, scene & distance between two entities (body parts, objects, surfaces, furniture) every 0.1\,s, with the direction at the closest moment\\
contacts & body & contacts of the body with the scene during the try\\
pose & body & pose that the body executed at a time of the try, in the format of a full-body keyframe\\
extent & scene & corners and heights of an object, a surface or a piece of furniture\\
fk & -- & joint positions of a pose written as a full-body keyframe, before it is used in a try (Sec.~\ref{sec:request})\\
\bottomrule
\end{tabular}
\caption{The queries of a measurement and the kind of quantitative feedback of Tab.~\ref{tab:feedback_kinds} that they give.}
\label{tab:supp_queries}
\end{table}

\begin{figure*}[tp]
\begin{agentbox}
generate_segment("intent": "pick up the toy airplane from the table with the right hand", "text": "A person takes a small step forward, reaches out with the right hand and picks up an object from the table in front of them, then stands holding it.", "seconds": 3, "constraints": [{"type": "root2d", "frame_indices": [40, 90], "smooth_root_2d": [[-0.05, 0.25], [-0.05, 0.25]], "global_root_heading": [[1, 0], [1, 0]]}], "events": [{"kind": "grasp", "hand": "right", "target": "072-a_toy_airplane_:0000", "t0": 0.5, "t1": 2.5}])
\end{agentbox}
\begin{resultbox}
Try 1 of segment 2 executed (not committed). Goal satisfied: no.
8 frames of the video of this try are attached (time stamps on each frame; in each frame left: camera following you; right: side camera). Use inspect to measure what happened.
\end{resultbox}
\begin{agentbox}
inspect("queries": [{"fn": "where", "id": "072-a_toy_airplane_:0000"}, {"fn": "contacts"}])
\end{agentbox}
\begin{resultbox}
[1] 072-a_toy_airplane_:0000 now: map (-1.28, 1.27), height 1.43; in your right hand; from your pelvis: forward +0.30, left -0.34 m; in Kimodo's frame of your next segment: X -0.47, Y 1.43, Z +0.41
[2] contacts: none
\end{resultbox}
\begin{agentbox}
commit()
\end{agentbox}
\begin{resultbox}
Committed segment 2.

# Current state (near you; get_state() gives the full state)
## You (the person)
- pelvis at map (x -0.99, y 0.91), facing heading 178 deg = west (map: +x = east = 0, +y = north = 90, -y = south = -90; turning left increases the heading)
- holding: right hand: 072-a_toy_airplane_:0000
[...]
\end{resultbox}
\caption{Excerpt of an episode of Inhabitant with Claude Opus 5.5 on the task ``Could you tidy up the toy airplane? It goes in the cardboard box by the TV stand.'' The agent requests a try for its second segment with a root waypoint and a grasp event, measures where the airplane is after the try, and commits the candidate segment. Blue boxes are the calls of the agent and grey boxes the results that it receives. The state description is cut.}
\label{fig:supp_example}
\end{figure*}

\subsection{Choice of motion generator}
\label{sec:supp_generator}

We compare Kimodo with OmniControl~\cite{xie2024omnicontrol}, a text-to-motion model with spatial control of joints, and with Kimodo given only text prompts, as a plain text-to-motion model.
The control specifications are those of the segments that Claude Opus 5.5 committed in the main runs.
Over all 526 committed segments, 8.7\% contain a full-body keyframe, 1.3\% several text prompts and 65.8\% a heading of the root, none of which OmniControl takes as input.
In addition, 85.7\% of the segments continue from the executed end of the previous one, while OmniControl starts every motion from its own canonical first frame.
To compare the generators on what both can express, we sample 120 committed segments without full-body keyframes or several text prompts, 60 with only root waypoints and 60 with end-effector targets.
Every control specification is generated once from a standing start with the same seed.
OmniControl receives the same text prompt and duration, the pelvis position on the ground for each root waypoint, and the 3D positions of the constrained joints for each end-effector target, computed with Kimodo's kinematics from the pose in the control specification.
We measure the errors at the constrained frames in the frame of the segment, after resampling OmniControl from 20 to 30 frames per second.

\begin{table}[h]
\centering
\small
\resizebox{\linewidth}{!}{%
\begin{tabular}{lccccc}
\toprule
Generator & Root (m) & End effector (m) & Within 10\,cm & Heading ($^\circ$) & Time (s)\\
\midrule
Kimodo, full control specification & 0.035 & 0.005 & 100\% & 1.5 & 3.7\\
OmniControl & 0.085 & 0.057 & 81\% & 39.9 & 95.4\\
Kimodo, text prompts only & 1.886 & 0.708 & 0\% & 52.9 & 3.6\\
\bottomrule
\end{tabular}}
\caption{Generators on 120 control specifications of the agent: mean error of the root at its waypoints (214 targets) and of the constrained joints of the end-effector targets (133 targets), share of end-effector targets reached within 10\,cm, mean heading error at the waypoints that set a heading, and generation time per segment on a V100. OmniControl has no heading input.}
\label{tab:supp_generator}
\end{table}

Kimodo reaches the end-effector targets almost exactly and the root waypoints within a few centimetres (Tab.~\ref{tab:supp_generator}).
OmniControl reaches the median root waypoint within 2\,cm, but one in ten misses by more than 25\,cm, and 81\% of its end-effector targets are within 10\,cm.
It cannot set the heading, which the agent uses in two thirds of its segments to face an object or a piece of furniture.
It also takes about 25 times as long per segment, which would add about half an hour of generation to an episode of 18 tries.
Text prompts alone miss the targets by metres, as with InteractAgent$^\dagger$ in the main paper.
Foot skate of the generated motion is higher for Kimodo (19.7\% of frames, 27.7\% in segments with only root waypoints) than for OmniControl (13.0\%) and for text prompts only (13.6\%), so we do not claim better motion quality for Kimodo.
The comparison has limits.
All motions start from a standing pose although the agent wrote the control specifications for the executed end of the previous segment, OmniControl's HumanML3D body differs from Kimodo's SMPL-X body by a few centimetres, and the motions are not executed in the simulator.

\section{Details of Inhabit}
\label{sec:supp_bench}
This section supplements Sec.~\ref{sec:bench}.

\subsection{Simulation}
\label{sec:supp_sim}

A grasp event succeeds if the palm centre, 5.4\,cm beyond the wrist, stays within 10\,cm of the object for 0.1\,s in its time window. A release event lets go of the object with the velocity of the hand, so that an object can be thrown. A drawer or a fridge door follows the hand that holds its handle.
The body is driven kinematically, without a balance or torque model, and execution is not exactly repeatable: identical inputs can differ by a few centimetres, and every execution is taken as it happened.

The executed body drifts backward.
On 286 tries of early development episodes we compared the pelvis of the motion as it is placed in the scene with the pelvis as executed.
In the 108 tries in which the body should stay in place, the executed pelvis ends a median of 8.6\,cm behind its target along the facing of the body, and the lag grows steadily at about 3\,cm/s (1.8\,cm after 0.5\,s, 6.4\,cm after 2\,s).
Both feet move back with the pelvis, so the whole body slides on the floor, and the drift is faster when the trunk leans forward (about 4\,cm/s).
The transfer to the simulated skeleton and the inverse kinematics contribute less than 2\,mm.
We keep the simulation as released and do not correct the drift: the agent measures where the body ended and compensates for it in its next try.

To quantify the repeatability, we take all pairs of consecutive tries in runs of Inhabitant with the five language models on all 75 tasks that start from the same state with an identical control specification and identical interaction events (257 pairs over the five language models).
The final pelvis positions of a pair differ by a median of 0.3\,cm, by less than 1\,cm in 81\% of the pairs and by less than 5\,cm in 93\%.

\subsection{The 75 tasks}
\label{sec:supp_tasks}
\begin{table}[h]
\centering
\small
\resizebox{\linewidth}{!}{%
\begin{tabular}{lrrrr}
\toprule
Tier & Tasks & with throw & by foot & EB original\\
\midrule
low (1 goal) & 30 & 5 & 5 & 3\\
mid (2 goals) & 20 & 3 & 4 & 3\\
high (3 goals) & 15 & 0 & 3 & 2\\
posture & 10 & -- & -- & 0\\
\midrule
total & 75 & 8 & 12 & 8\\
\bottomrule
\end{tabular}}
\caption{Composition of the tasks: number of tasks that contain a throw goal or a goal of moving an object with a foot, and original EB-Habitat episodes. All other goals are place goals. The 10 posture tasks are sitting (4), lying (2) and kneeling (4). Goal counts of EB originals follow EB's own goal.}
\label{tab:tasks}
\end{table}
A throw goal counts only if the object is released with the wrist at least 1\,m (horizontally) from the target region, and a goal of moving an object with a foot only if the object was never grasped.
Goals of EB-Habitat's 300 episodes need only grasp and release events, whereas ours also require throwing, kicking, pushing and postures.

Tab.~\ref{tab:supp_conditions} defines the conditions of the success rule, and Tabs.~\ref{tab:supp_tasks_low} to \ref{tab:supp_tasks_high} list the instruction and the conditions of every task.
The 20 tasks of the ablation set, marked in these tables, cover all four tiers (9 low, 4 mid, 3 high and 4 posture tasks) and all five goal types.

\begin{table}[h]
\centering
\small
\setlength{\tabcolsep}{3pt}
\begin{tabular}{@{}lp{0.8\linewidth}@{}}
\toprule
Condition & Holds if\\
\midrule
on & the object rests inside the region of the target surface and is not held\\
in & the centre of the object is inside the target box or basket, wherever that container is now, and the object is not held\\
inside & the object is inside the fridge and is not held\\
open & the fridge door is opened by more than 1.07\,rad, or the drawer is pulled out by more than 0.35\,m\\
closed & the joint of the door or drawer is within 0.15 of its closed position\\
throw & the object ends on or in the target, and the palm was at least 1\,m (horizontally) from the target when the object left the hand\\
kick & the object ends at the target after a foot touched it, and it is not in a hand from that contact on\\
push & the object ends at the target, was never held, and the body touched it\\
sit & for 1\,s, the pelvis is over the surface and 0.05 to 0.35\,m above it, with the trunk within 45$^\circ$ of vertical\\
lie & for 1\,s, the trunk is more than 60$^\circ$ from vertical, with the pelvis and the head over the surface and the pelvis at most 0.35\,m above it\\
kneel & for 1\,s, a knee is on the floor and the pelvis is 0.25 to 0.8\,m above it\\
avoid & the body never touches the named object\\
together & the two objects are held at the same time, one in each hand\\
EB & the goal of the original EB-Habitat episode holds\\
\bottomrule
\end{tabular}
\caption{The conditions of the success rules. The target of a kick or a push can be the floor under a piece of furniture, where the object must lie less than 0.15\,m above the floor.}
\label{tab:supp_conditions}
\end{table}

\begin{table*}[tp]
\centering
\scriptsize
\setlength{\tabcolsep}{4pt}
\begin{tabular}{@{}lp{0.68\linewidth}p{0.2\linewidth}@{}}
\toprule
Task & Instruction & Conditions\\
\midrule
low\_01 & Could you tidy up the toy airplane? It goes in the cardboard box by the TV stand. & in\\
low\_02 & Put the lego in the cardboard box next to the black table. & in\\
low\_03 & Put the toy airplane in the cardboard box next to the black table. & in\\
low\_04 & The towel is on the floor. Put it back on the sofa. & on\\
low\_05 & There's a towel lying on the floor - could you put it back on the sofa? & on\\
low\_06 & Pick up the remote control from the floor and put it on the TV stand. & on\\
low\_07$^\ast$ & Could you put the can away on the top shelf of the tall shelf next to the right counter? & on\\
low\_08 & Put the ball on the top shelf of the tall shelf next to the brown table. & on\\
low\_09 & Could you put the orange away on the bottom shelf of the TV stand? & on\\
low\_10 & Please bring the dirty coffee mug on the brown table to the sink. & on\\
low\_11$^\ast$ & Bring the coffee mug on the brown table to the sink - careful, don't touch the vase beside it. & on, avoid\\
low\_12$^\ast$ & Put the banana in the fridge. & inside\\
low\_13 & Put the apple in the fridge. & inside\\
low\_14 & Please open the fridge. & open\\
low\_15 & The fridge door is open, please close it. & closed\\
low\_16$^\ast$ & Could you pull the middle drawer in the kitchen open? & open\\
low\_17 & Open the right drawer of the kitchen counter. & open\\
low\_18$^\ast$ & The fridge door is open because of my mistake. Can you close it, please? & EB\\
low\_19 & Hey, on the sink, I accidentally left my plate, can you bring it to the right counter? & EB\\
low\_20 & Can you bring something to assist in cutting paper at the sink? & EB\\
low\_21$^\ast$ & Toss the ball into the cardboard box by the TV stand from where you can, at least a metre away. & throw\\
low\_22 & Toss the toy airplane into the basket by the TV stand from where you can, at least a metre away. & throw\\
low\_23 & Throw the ball lying on the floor onto the sofa - from a metre or more. & throw\\
low\_24$^\ast$ & Toss the cushion onto the sofa from at least a metre away. & throw\\
low\_25 & Throw the towel lying on the floor onto the sofa - from a metre or more. & throw\\
low\_26$^\ast$ & There's a ball on the floor - kick it under the black table. & kick\\
low\_27 & Kick the ball on the floor under the black table. & kick\\
low\_28 & Nudge the shoe box under the black table - no need to pick it up, just push it with your foot. & push\\
low\_29 & Nudge the cardboard box under the black table - no need to pick it up, just push it with your foot. & push\\
low\_30$^\ast$ & Push the shoe on the floor out of the way with your foot (half a metre or more); don't pick it up. & push\\
\bottomrule
\end{tabular}
\caption{The 30 low tasks of Inhabit: instruction and conditions of the success rule (Tab.~\ref{tab:supp_conditions}). $^\ast$ marks a task of the ablation set.}
\label{tab:supp_tasks_low}
\end{table*}

\begin{table*}[tp]
\centering
\scriptsize
\setlength{\tabcolsep}{4pt}
\begin{tabular}{@{}lp{0.68\linewidth}p{0.2\linewidth}@{}}
\toprule
Task & Instruction & Conditions\\
\midrule
mid\_01 & Could you tidy the toy airplane and the rubriks cube into the basket by the black table? & 2$\times$in\\
mid\_02 & Could you tidy the lego and the toy airplane into the cardboard box by the sofa? & 2$\times$in\\
mid\_03 & Tidy up: the rubriks cube goes in the basket next to the sofa; throw the toy airplane in from a metre or more. & in, throw\\
mid\_04 & The fridge is open. Put the banana inside and close the door. & inside, closed (ordered)\\
mid\_05 & The fridge is open. Put the peach inside and close the door. & inside, closed (ordered)\\
mid\_06$^\ast$ & Take both the coffee mug and the serving bowl on the black table to the sink in one go - one in each hand. & 2$\times$on, together\\
mid\_07 & Take both the coffee mug and the serving bowl on the brown table to the sink in one go - one in each hand. & 2$\times$on, together\\
mid\_08 & Tidy the rubriks cube into the cardboard box next to the sofa, and push the shoe on the floor out of the way with your foot (at least half a metre). & in, push\\
mid\_09 & Put the lego in the cardboard box by the sofa. There's a shoe lying in the way - nudge it aside with your foot, half a metre or more, without picking it up. & in, push\\
mid\_10 & Clear the coffee mug and the dinner plate from the brown table into the sink. & 2$\times$on\\
mid\_11 & Please take the dirty coffee mug and dinner plate on the black table to the sink - without touching the vase next to them. & 2$\times$on, avoid\\
mid\_12 & Put the ball on the top shelf of the tall shelf next to the black table and the orange on the bottom shelf of the TV stand. & 2$\times$on\\
mid\_13 & The cushion and the towel are on the floor: put the cushion back on the sofa and throw the towel onto it from a metre or more. & on, throw\\
mid\_14 & Two things are on the floor: put the cushion back on the sofa, and toss the towel onto the sofa from at least a metre away. & on, throw\\
mid\_15$^\ast$ & Tidy up: the rubriks cube goes in the cardboard box next to the sofa, and the ball gets kicked back under the black table. & in, kick\\
mid\_16 & Please tidy up: put the rubriks cube in the cardboard box and kick the ball back under the black table. & in, kick\\
mid\_17$^\ast$ & Could you put the banana away in the left drawer in the kitchen? & open, inside (ordered)\\
mid\_18$^\ast$ & Transfer all ball to the brown table. & EB\\
mid\_19 & Transfer all lego to the sofa. & EB\\
mid\_20 & Transport all plate and put them on the right counter. & EB\\
\bottomrule
\end{tabular}
\caption{The 20 mid tasks of Inhabit, as in Tab.~\ref{tab:supp_tasks_low}. ``Ordered'' means that the conditions must become true in the order of the instruction.}
\label{tab:supp_tasks_mid}
\end{table*}

\begin{table*}[tp]
\centering
\scriptsize
\setlength{\tabcolsep}{4pt}
\begin{tabular}{@{}lp{0.68\linewidth}p{0.2\linewidth}@{}}
\toprule
Task & Instruction & Conditions\\
\midrule
high\_01 & Clear the sofa: all the things on it go on the top shelf of the tall shelf next to the black table. & 3$\times$on\\
high\_02 & Clear the sofa: all the things on it go on the top shelf of the tall shelf next to the brown table. & 3$\times$on\\
high\_03$^\ast$ & Take everything out of the fridge, put it on the black table, and close the fridge again. & 2$\times$on, closed\\
high\_04 & Empty the fridge onto the black table, then shut the fridge door. & 2$\times$on, closed\\
high\_05 & Put all the things on the black table into the cardboard box beside it. & 3$\times$in\\
high\_06 & Clear everything off the black table into the cardboard box next to it. & 3$\times$in\\
high\_07$^\ast$ & Could you tidy up? Put the lego in the basket next to the sofa, kick the ball under the black table, and the cushion on the floor goes back on the sofa. & in, kick, on\\
high\_08 & Could you tidy up? Put the rubriks cube in the basket next to the sofa, kick the ball under the black table, and the cushion on the floor goes back on the sofa. & in, kick, on\\
high\_09 & Could you tidy up? Put the toy airplane in the basket next to the TV stand, kick the ball under the black table, and the cushion on the floor goes back on the sofa. & in, kick, on\\
high\_10$^\ast$ & Please move the coffee mug, dinner plate and serving bowl on the right counter to the black table. & 3$\times$on\\
high\_11 & Set the black table: bring the coffee mug, the dinner plate and the serving bowl from the right counter. & 3$\times$on\\
high\_12 & Put the apple and the plum in the fridge and close the door afterwards. & 2$\times$inside, closed (ordered)\\
high\_13 & Put the peach and the lemon in the fridge and close the door afterwards. & 2$\times$inside, closed (ordered)\\
high\_14 & Move the ball to the brown table, the block to the brown table, and the hammer to the black table. & EB\\
high\_15 & Move the lemon to the left counter, the block to the brown table, and the bowl to the brown table. & EB\\
posture\_01$^\ast$ & Pick the cushion up off the floor and put it on the sofa; after that, take a seat on the sofa. & on, sit (ordered)\\
posture\_02 & Put the cushion from the floor back on the sofa, then sit down on the sofa for a moment. & on, sit (ordered)\\
posture\_03$^\ast$ & Pick the cushion up off the floor and put it on the sofa; after that, have a rest lying on the sofa. & on, lie (ordered)\\
posture\_04 & Put the cushion from the floor back on the sofa, then lie down on the sofa for a moment. & on, lie (ordered)\\
posture\_05$^\ast$ & Bring the coffee mug from the right counter to the black table and sit down on the chair. & on, sit (ordered)\\
posture\_06 & Take the coffee mug on the right counter over to the black table, then have a seat on the chair. & on, sit (ordered)\\
posture\_07 & Get down on your knees and put the cup away on the bottom shelf of the TV stand. & on, kneel\\
posture\_08 & Get down on your knees and put the apple away on the bottom shelf of the TV stand. & on, kneel\\
posture\_09$^\ast$ & The toy airplane rolled under the black table. Kneel down, get it out and put it in the cardboard box. & in, kneel\\
posture\_10 & The block rolled under the black table. Kneel down, get it out and put it in the cardboard box. & in, kneel\\
\bottomrule
\end{tabular}
\caption{The 15 high tasks and the 10 posture tasks of Inhabit, as in Tab.~\ref{tab:supp_tasks_low}.}
\label{tab:supp_tasks_high}
\end{table*}

\subsection{Metrics}
\label{sec:supp_metrics}

\emph{Success} is the share of episodes whose final state satisfies the success rule, and \emph{sub-goal} the mean share of the conditions of the success rule that hold at the end, where a required order between two conditions counts as a condition of its own.
\emph{Collisions} count the contacts of the body with furniture, walls and pillars in the sequence, excluding the feet on the floor and the hands on a grasped object. \emph{Disturbed objects} are objects that the task does not ask to move and that end more than 5\,cm from their start.
\emph{Skate} is the foot-skate ratio of Karunratanakul~\etal~\cite{karunratanakul2023guided}. It partly reflects a backward drift of the body during execution (about 3\,cm/s while standing), so we use it only as supporting evidence.
\emph{Tries} count the tries of an episode, \emph{calls} the calls to the language model, and \emph{tokens} the size of the history at the end of the episode, which is comparable across language models.
In the ablations we additionally report \emph{missed grasps}, which take every grasp event of every try of an episode and count the share that does not attach the object, that is, where the palm centre does not stay within 10\,cm of the object for 0.1\,s in the time window. The share is computed per episode and averaged over the episodes, and an episode without grasp events counts as 0.

\subsection{Corrections of success rules}
\label{sec:supp_corrections}
After the main runs we corrected three success rules and re-judged all runs of every model, baseline and ablation from their recorded final states, without running anything again. The agents never saw any of these rules, since every try only reports whether the success rule is satisfied.
(1)~high\_15, an original EB-Habitat task, asks for a bowl. Our benchmark renames the ReplicaCAD bowl to ``serving bowl'' in its text, and the same name was used when EB's goal looked for an object of category bowl, so the goal could never be met. EB's goals now use EB's own categories. Opus 5.5 had completed the task.
(2)~high\_03 and high\_04 required both fruits to be on the table before the fridge door closes. Closing the door while still holding the last fruit and then putting it down also completes the instruction, so only the final state is checked now. This changes high\_03 of the Single-Try ablation.
(3)~Kneeling required the trunk to be within 30 degrees of vertical. Kneeling and leaning under a table to fetch a toy failed that test, so a kneel now needs a knee on the floor with the pelvis between 0.25 and 0.8\,m above it for one second. This changes body\_09 of gpt-6-luna.
The records are in the released runs (runs/checks/rejudge\_*).

\section{Experimental setup}
\label{sec:supp_setup}
This section describes how the experiments of Sec.~\ref{sec:exp_setup} are carried out.

\subsection{Implementation}
\label{sec:supp_impl}
\paragraph{Language models.}
Claude Opus 5.5 and Claude Sonnet 5.5 use adaptive thinking at medium effort, Gemini 4 Argon a medium thinking level at temperature 1, gpt-6-luna (version 2026-09-22) medium reasoning effort, and GLM-5.3-Flash high reasoning effort at temperature 1 and top-$p$ 0.95.
Every language model receives the same system prompt and the same tools (Supp.~\ref{sec:supp_prompts}).
The history of an episode is never truncated or summarized, so the rendered frames of all tries stay in it; the eight frames of a try are sent as one image, 768 pixels wide.

\paragraph{Motion generator.}
We use the released Kimodo model for the SMPL-X skeleton (Kimodo-SMPLX-RP-v1) without any training, at 30 frames per second.
A segment lasts 0.5 to 8\,s.
Unless the agent sets them, the seed is 0, post-processing is on, and the guidance weights of the text and of the constraints are both 2, the defaults of Kimodo.
After the motion of a segment the simulation runs on for 0.3\,s, or 1.5\,s after a release event, so that the objects settle.

\paragraph{Budget and run time.}
An episode has a budget of 150 calls to the language model, and every task is run once.
One NVIDIA V100 GPU hosts the motion generator and the simulations of several episodes.
A try of Claude Opus 5.5 takes about 45\,s for generation, execution and rendering, and an episode 21 minutes on average, of which 6 are spent in the language model.
The episodes of the other language models take longer: 38 minutes for Claude Sonnet 5.5, 42 for Gemini 4 Argon, 90 for gpt-6-luna and 110 for GLM-5.3-Flash.

\paragraph{Ablations.}
An ablation changes the prompt only where it describes the component that is turned off.
For the granularity of spatial control (Tab.~\ref{tab:control}), one sentence states which types of constraints are available, and constraints of the other types are rejected; with text prompts only, the constraints and their documentation are removed.
For the granularity of feedback (Tab.~\ref{tab:feedback}), the description of the measurements lists only the queries of the remaining kind, or the measurement tool is removed.
Without the success check, a try no longer reports whether the success rule is satisfied, and the agent ends the episode with an additional call when it considers the task done.
In Single-Try the commit and the rollback are removed, and in Open-Loop the agent has a single call that takes the control specifications of all segments.

\subsection{Adapted baselines}
\label{sec:supp_baselines}
The three baselines of the main paper are adaptations and do not run the original systems. Each keeps the control interface of the original agent, and all use Claude Opus 5.5, our scene simulator, the state description, the interaction events, the success rule and the budget of 150 calls.
InteractAgent$^\dagger$ follows the loop of InteractAgent with a text-to-motion generator (MotionLLM): the agent writes only text prompts, and the generator takes no duration, seed or spatial constraint. After each try the agent receives the eight frames, the success check and the state description, and it has no measurement queries. It first writes a plan of steps, may redo each step twice before the segment is committed, and may restart the episode from the beginning up to five times. It differs from the setting with text prompts only of Tab.~\ref{tab:control}, which keeps our motion generator and all our feedback.
HumanCLAW$^\dagger$ calls the skills of HumanCLAW (walk, turn, side step, step back, sit and stop). Each skill is executed at once and cannot be revised, and the agent has no measurement queries. It sees the rendered frames and the state description after each skill.
GEMS$^\dagger$ may query positions and extents of the initial scene, then submits one plan of segments with text prompts and full-body keyframes for the same motion generator. The plan is executed once and the agent receives no feedback on it.

\section{Additional results}
\label{sec:supp_results}
This section gives the details of the results of Secs.~\ref{sec:main} to \ref{sec:loop_exp}, in the order of the main paper.

\subsection{Baselines and language models (Sec.~\ref{sec:main})}
\label{sec:supp_main}
Tab.~\ref{tab:supp_failures} shows how the failed episodes of each baseline end and how many of its grasps miss, which supports the three failure modes of Sec.~\ref{sec:main}.
InteractAgent$^\dagger$ fails on 55 tasks, and 46 of these episodes use up the budget of 150 calls; in the other nine the agent stops by itself after 101 to 149 calls.
HumanCLAW$^\dagger$ fails on 60 tasks, and in every one of them the agent stops by itself, after 12 calls on average.
GEMS$^\dagger$ executes its single plan to the end in every episode and fails in 70 of them; four of its five successes contain a throw or a push, whose targets are loose.
All three miss most of their grasps, against 48\% for Inhabitant with the same language model.

\begin{table}[h]
\centering
\small
\setlength{\tabcolsep}{3.5pt}
\resizebox{\linewidth}{!}{%
\begin{tabular}{lccccc}
\toprule
 & Failed & Budget & Stopped & Plan & Missed\\
Baseline & episodes & used up & by the agent & executed & grasps\\
\midrule
InteractAgent$^\dagger$ & 55 & 46 & 9 & -- & 83\%\\
HumanCLAW$^\dagger$ & 60 & 0 & 60 & -- & 75\%\\
GEMS$^\dagger$ & 70 & -- & -- & 70 & 68\%\\
\bottomrule
\end{tabular}}
\caption{How the failed episodes of the baselines of Tab.~\ref{tab:main} end, and their missed grasps over all 75 episodes (Supp.~\ref{sec:supp_metrics}). GEMS$^\dagger$ has one call for its plan, so its episodes end when the plan is executed.}
\label{tab:supp_failures}
\end{table}

Tab.~\ref{tab:supp_main_pertask} gives the result of every task for the eight runs of Tab.~\ref{tab:main}.
No baseline solves a high task except HumanCLAW$^\dagger$ on one, and 48 of the 75 tasks are solved by none of the three baselines.

Tab.~\ref{tab:supp_cost} gives the cost of the runs of Tab.~\ref{tab:main} in API fees.
For every call of the language model, we take the tokens that the provider bills (input, input read from the prompt cache, and output including reasoning) and multiply them by its list prices.
With Claude Opus 5.5, an episode of Inhabitant costs \$1.70 and a solved task \$1.72.
The baselines with the same language model cost \$0.96 to \$2.56 per episode but solve few tasks, so a solved task costs \$4.80 to \$21.08.
Among the three stronger language models, Claude Sonnet 5.5 is the cheapest per solved task (\$1.26).
gpt-6-luna costs less per solved task (\$0.82) because its list price is forty times lower, but it solves only 23 tasks.
GLM-5.3-Flash has open weights and no list price.
The GPU that runs the motion generator and the scene simulator is not included; one V100 is enough for both.

\begin{table}[h]
\centering
\small
\setlength{\tabcolsep}{3.5pt}
\resizebox{\linewidth}{!}{%
\begin{tabular}{llccccc}
\toprule
 & & \multicolumn{3}{c}{Tokens per episode} & \multicolumn{2}{c}{Cost (USD)}\\
\cmidrule(lr){3-5}\cmidrule(lr){6-7}
Method & Language model & Input & Cached & Output & Episode & Solved task\\
\midrule
InteractAgent$^\dagger$ & Claude Opus 5.5 & 6.3M & 98\% & 39k & 2.56 & 9.62\\
HumanCLAW$^\dagger$ & Claude Opus 5.5 & 1.5M & 95\% & 16k & 0.96 & 4.80\\
GEMS$^\dagger$ & Claude Opus 5.5 & 1.3M & 92\% & 32k & 1.41 & 21.08\\
\midrule
\multirow{5}{*}{Inhabitant} & GLM-5.3-Flash & 15.4M & 0\% & 106k & -- & --\\
 & gpt-6-luna & 10.5M & 87\% & 51k & 0.25 & 0.82\\
 & Gemini 4 Argon & 5.3M & 89\% & 31k & 1.94 & 2.35\\
 & Claude Sonnet 5.5 & 5.0M & 98\% & 42k & 1.18 & 1.26\\
 & Claude Opus 5.5 & 3.2M & 97\% & 30k & 1.70 & 1.72\\
\bottomrule
\end{tabular}}
\caption{Cost in API fees of the runs of Tab.~\ref{tab:main}.}
\label{tab:supp_cost}
\end{table}

\begin{table*}[tp]
\centering
\scriptsize
\setlength{\tabcolsep}{2.6pt}
\begin{tabular}{@{}lccccccccc@{\hspace{0.6em}}lcccccccc@{}}
\toprule
Task & IA$^\dagger$ & HC$^\dagger$ & GEMS$^\dagger$ & GLM & luna & Gemini & Sonnet & Opus & & Task & IA$^\dagger$ & HC$^\dagger$ & GEMS$^\dagger$ & GLM & luna & Gemini & Sonnet & Opus\\
\midrule
low\_01 & \checkmark & \checkmark & $\cdot$ & \checkmark & $\cdot$ & \checkmark & \checkmark & \checkmark & & mid\_09 & $\cdot$ & $\cdot$ & $\cdot$ & $\cdot$ & \checkmark & \checkmark & \checkmark & \checkmark\\
low\_02 & \checkmark & $\cdot$ & $\cdot$ & \checkmark & $\cdot$ & \checkmark & \checkmark & \checkmark & & mid\_10 & $\cdot$ & $\cdot$ & $\cdot$ & $\cdot$ & $\cdot$ & \checkmark & \checkmark & \checkmark\\
low\_03 & $\cdot$ & $\cdot$ & $\cdot$ & \checkmark & $\cdot$ & \checkmark & \checkmark & \checkmark & & mid\_11 & $\cdot$ & $\cdot$ & $\cdot$ & $\cdot$ & $\cdot$ & \checkmark & \checkmark & \checkmark\\
low\_04 & \checkmark & \checkmark & \checkmark & \checkmark & \checkmark & \checkmark & \checkmark & \checkmark & & mid\_12 & $\cdot$ & $\cdot$ & $\cdot$ & $\cdot$ & $\cdot$ & $\cdot$ & \checkmark & \checkmark\\
low\_05 & \checkmark & \checkmark & $\cdot$ & \checkmark & \checkmark & \checkmark & \checkmark & \checkmark & & mid\_13 & $\cdot$ & $\cdot$ & \checkmark & $\cdot$ & $\cdot$ & \checkmark & \checkmark & \checkmark\\
low\_06 & $\cdot$ & \checkmark & $\cdot$ & \checkmark & \checkmark & \checkmark & \checkmark & \checkmark & & mid\_14 & $\cdot$ & $\cdot$ & $\cdot$ & $\cdot$ & $\cdot$ & \checkmark & \checkmark & \checkmark\\
low\_07$^\ast$ & $\cdot$ & $\cdot$ & $\cdot$ & $\cdot$ & $\cdot$ & \checkmark & \checkmark & \checkmark & & mid\_15$^\ast$ & $\cdot$ & $\cdot$ & $\cdot$ & $\cdot$ & $\cdot$ & \checkmark & \checkmark & \checkmark\\
low\_08 & \checkmark & $\cdot$ & $\cdot$ & \checkmark & $\cdot$ & \checkmark & \checkmark & \checkmark & & mid\_16 & $\cdot$ & \checkmark & $\cdot$ & $\cdot$ & $\cdot$ & \checkmark & \checkmark & \checkmark\\
low\_09 & $\cdot$ & $\cdot$ & $\cdot$ & $\cdot$ & $\cdot$ & \checkmark & \checkmark & \checkmark & & mid\_17$^\ast$ & $\cdot$ & \checkmark & $\cdot$ & \checkmark & \checkmark & \checkmark & \checkmark & \checkmark\\
low\_10 & $\cdot$ & $\cdot$ & $\cdot$ & \checkmark & \checkmark & \checkmark & \checkmark & \checkmark & & mid\_18$^\ast$ & $\cdot$ & $\cdot$ & $\cdot$ & $\cdot$ & \checkmark & \checkmark & \checkmark & \checkmark\\
low\_11$^\ast$ & \checkmark & $\cdot$ & $\cdot$ & \checkmark & $\cdot$ & \checkmark & \checkmark & \checkmark & & mid\_19 & $\cdot$ & $\cdot$ & $\cdot$ & $\cdot$ & $\cdot$ & \checkmark & $\cdot$ & \checkmark\\
low\_12$^\ast$ & \checkmark & $\cdot$ & $\cdot$ & \checkmark & $\cdot$ & \checkmark & \checkmark & \checkmark & & mid\_20 & $\cdot$ & $\cdot$ & $\cdot$ & $\cdot$ & \checkmark & \checkmark & \checkmark & \checkmark\\
low\_13 & $\cdot$ & $\cdot$ & $\cdot$ & $\cdot$ & $\cdot$ & \checkmark & \checkmark & \checkmark & & high\_01 & $\cdot$ & $\cdot$ & $\cdot$ & $\cdot$ & $\cdot$ & $\cdot$ & \checkmark & \checkmark\\
low\_14 & \checkmark & $\cdot$ & $\cdot$ & \checkmark & \checkmark & \checkmark & \checkmark & \checkmark & & high\_02 & $\cdot$ & $\cdot$ & $\cdot$ & $\cdot$ & $\cdot$ & \checkmark & $\cdot$ & \checkmark\\
low\_15 & \checkmark & $\cdot$ & $\cdot$ & \checkmark & \checkmark & \checkmark & \checkmark & \checkmark & & high\_03$^\ast$ & $\cdot$ & $\cdot$ & $\cdot$ & $\cdot$ & $\cdot$ & $\cdot$ & \checkmark & \checkmark\\
low\_16$^\ast$ & \checkmark & \checkmark & $\cdot$ & \checkmark & \checkmark & \checkmark & \checkmark & \checkmark & & high\_04 & $\cdot$ & $\cdot$ & $\cdot$ & $\cdot$ & $\cdot$ & \checkmark & \checkmark & \checkmark\\
low\_17 & \checkmark & \checkmark & $\cdot$ & \checkmark & \checkmark & \checkmark & \checkmark & \checkmark & & high\_05 & $\cdot$ & $\cdot$ & $\cdot$ & $\cdot$ & $\cdot$ & \checkmark & \checkmark & \checkmark\\
low\_18$^\ast$ & \checkmark & $\cdot$ & $\cdot$ & \checkmark & $\cdot$ & \checkmark & \checkmark & \checkmark & & high\_06 & $\cdot$ & $\cdot$ & $\cdot$ & $\cdot$ & $\cdot$ & $\cdot$ & \checkmark & \checkmark\\
low\_19 & \checkmark & $\cdot$ & $\cdot$ & \checkmark & $\cdot$ & \checkmark & $\cdot$ & \checkmark & & high\_07$^\ast$ & $\cdot$ & $\cdot$ & $\cdot$ & $\cdot$ & $\cdot$ & \checkmark & \checkmark & \checkmark\\
low\_20 & $\cdot$ & $\cdot$ & $\cdot$ & $\cdot$ & $\cdot$ & \checkmark & \checkmark & \checkmark & & high\_08 & $\cdot$ & \checkmark & $\cdot$ & $\cdot$ & $\cdot$ & \checkmark & $\cdot$ & \checkmark\\
low\_21$^\ast$ & $\cdot$ & $\cdot$ & $\cdot$ & $\cdot$ & $\cdot$ & \checkmark & \checkmark & \checkmark & & high\_09 & $\cdot$ & $\cdot$ & $\cdot$ & $\cdot$ & $\cdot$ & $\cdot$ & \checkmark & \checkmark\\
low\_22 & $\cdot$ & $\cdot$ & $\cdot$ & \checkmark & $\cdot$ & \checkmark & \checkmark & \checkmark & & high\_10$^\ast$ & $\cdot$ & $\cdot$ & $\cdot$ & $\cdot$ & $\cdot$ & \checkmark & \checkmark & \checkmark\\
low\_23 & \checkmark & $\cdot$ & \checkmark & \checkmark & \checkmark & \checkmark & \checkmark & \checkmark & & high\_11 & $\cdot$ & $\cdot$ & $\cdot$ & $\cdot$ & $\cdot$ & \checkmark & \checkmark & \checkmark\\
low\_24$^\ast$ & $\cdot$ & $\cdot$ & $\cdot$ & \checkmark & \checkmark & \checkmark & \checkmark & \checkmark & & high\_12 & $\cdot$ & $\cdot$ & $\cdot$ & $\cdot$ & $\cdot$ & $\cdot$ & \checkmark & \checkmark\\
low\_25 & $\cdot$ & $\cdot$ & $\cdot$ & $\cdot$ & $\cdot$ & \checkmark & \checkmark & \checkmark & & high\_13 & $\cdot$ & $\cdot$ & $\cdot$ & $\cdot$ & $\cdot$ & $\cdot$ & \checkmark & \checkmark\\
low\_26$^\ast$ & $\cdot$ & $\cdot$ & $\cdot$ & \checkmark & $\cdot$ & \checkmark & \checkmark & \checkmark & & high\_14 & $\cdot$ & $\cdot$ & $\cdot$ & $\cdot$ & \checkmark & $\cdot$ & \checkmark & \checkmark\\
low\_27 & $\cdot$ & \checkmark & $\cdot$ & \checkmark & $\cdot$ & \checkmark & \checkmark & \checkmark & & high\_15 & $\cdot$ & $\cdot$ & $\cdot$ & $\cdot$ & $\cdot$ & $\cdot$ & \checkmark & \checkmark\\
low\_28 & $\cdot$ & \checkmark & \checkmark & \checkmark & $\cdot$ & \checkmark & \checkmark & \checkmark & & posture\_01$^\ast$ & \checkmark & \checkmark & $\cdot$ & \checkmark & \checkmark & \checkmark & \checkmark & \checkmark\\
low\_29 & \checkmark & \checkmark & $\cdot$ & \checkmark & \checkmark & \checkmark & \checkmark & \checkmark & & posture\_02 & \checkmark & $\cdot$ & $\cdot$ & \checkmark & $\cdot$ & \checkmark & \checkmark & \checkmark\\
low\_30$^\ast$ & \checkmark & \checkmark & \checkmark & \checkmark & \checkmark & \checkmark & \checkmark & \checkmark & & posture\_03$^\ast$ & $\cdot$ & $\cdot$ & $\cdot$ & \checkmark & \checkmark & \checkmark & \checkmark & $\cdot$\\
mid\_01 & $\cdot$ & $\cdot$ & $\cdot$ & $\cdot$ & $\cdot$ & \checkmark & \checkmark & \checkmark & & posture\_04 & $\cdot$ & $\cdot$ & $\cdot$ & \checkmark & \checkmark & \checkmark & \checkmark & \checkmark\\
mid\_02 & $\cdot$ & $\cdot$ & $\cdot$ & $\cdot$ & $\cdot$ & \checkmark & \checkmark & \checkmark & & posture\_05$^\ast$ & $\cdot$ & $\cdot$ & $\cdot$ & \checkmark & $\cdot$ & \checkmark & \checkmark & \checkmark\\
mid\_03 & $\cdot$ & $\cdot$ & $\cdot$ & $\cdot$ & $\cdot$ & $\cdot$ & $\cdot$ & \checkmark & & posture\_06 & $\cdot$ & $\cdot$ & $\cdot$ & \checkmark & \checkmark & \checkmark & \checkmark & \checkmark\\
mid\_04 & $\cdot$ & $\cdot$ & $\cdot$ & $\cdot$ & $\cdot$ & \checkmark & \checkmark & \checkmark & & posture\_07 & $\cdot$ & $\cdot$ & $\cdot$ & $\cdot$ & $\cdot$ & \checkmark & \checkmark & \checkmark\\
mid\_05 & \checkmark & $\cdot$ & $\cdot$ & $\cdot$ & $\cdot$ & $\cdot$ & \checkmark & \checkmark & & posture\_08 & $\cdot$ & $\cdot$ & $\cdot$ & $\cdot$ & $\cdot$ & $\cdot$ & \checkmark & \checkmark\\
mid\_06$^\ast$ & $\cdot$ & $\cdot$ & $\cdot$ & $\cdot$ & $\cdot$ & \checkmark & \checkmark & \checkmark & & posture\_09$^\ast$ & $\cdot$ & $\cdot$ & $\cdot$ & \checkmark & \checkmark & \checkmark & \checkmark & \checkmark\\
mid\_07 & $\cdot$ & $\cdot$ & $\cdot$ & $\cdot$ & $\cdot$ & \checkmark & \checkmark & \checkmark & & posture\_10 & $\cdot$ & $\cdot$ & $\cdot$ & $\cdot$ & $\cdot$ & $\cdot$ & \checkmark & \checkmark\\
mid\_08 & \checkmark & \checkmark & $\cdot$ & \checkmark & \checkmark & \checkmark & \checkmark & \checkmark & & \textbf{solved} & 20 & 15 & 5 & 33 & 23 & 62 & 70 & 74\\
\bottomrule
\end{tabular}
\caption{Per-task results of the runs of Tab.~\ref{tab:main}: \checkmark\ marks a success. IA$^\dagger$, HC$^\dagger$ and GEMS$^\dagger$ are the three baselines with Claude Opus 5.5, and the other columns are Inhabitant with the five language models. $^\ast$ marks a task of the ablation set. The last row counts the solved tasks out of 75.}
\label{tab:supp_main_pertask}
\end{table*}

\subsection{How the agent uses control and feedback (Secs.~\ref{sec:main} and \ref{sec:control_exp})}
\label{sec:supp_usage}

\begin{table*}[tp]
\centering
\small
\setlength{\tabcolsep}{3.4pt}
\begin{tabular}{lcccccc}
\toprule
 & Opus 5.5 & Sonnet 5.5 & Gemini & GLM & luna & InteractAgent$^\dagger$\\
\midrule
Segments with a spatial constraint & 97\% & 96\% & 100\% & 90\% & 90\% & --\\
\quad with an end-effector target or a full-body keyframe & 37\% & 26\% & 49\% & 17\% & 4\% & --\\
Committed at the first try & 46\% & 52\% & 48\% & 46\% & 36\% & 32\%\\
Tries followed by a measurement & 93\% & 99\% & 98\% & 80\% & 95\% & --\\
Revisions that change only the seed & 6\% & 7\% & 0\% & 1\% & 1\% & --\\
Missed grasps & 48\% & 66\% & 23\% & 73\% & 79\% & 83\%\\
\midrule
Next grasp succeeds after a miss, & & & & & & \\
\quad end-effector target or keyframe revised & 47\% & 27\% & 44\% & 24\% & 24\% & --\\
\quad end-effector target or keyframe unchanged & 21\% & 24\% & n/a & 9\% & 10\% & 4\%\\
\midrule
Failed episodes & 1 & 5 & 13 & 42 & 52 & 55\\
\quad budget used up & 0 & 2 & 11 & 40 & 45 & 46\\
\quad ended before the budget is used up & 1 & 3 & 2 & 2 & 7 & 9\\
\bottomrule
\end{tabular}
\caption{How the agent uses the spatial constraints and the feedback in the episodes of the main table, for Inhabitant with the five language models and for InteractAgent$^\dagger$, which has text prompts only and neither measurement queries nor a seed. Revisions are pairs of consecutive tries for the same segment. The rows on grasps take every try whose grasp failed and ask whether the grasp of the next try for the same segment succeeds, split by whether the agent changed an end-effector target or a full-body keyframe in between. The last rows show how the failed episodes end. --: does not apply; n/a: no such case.}
\label{tab:analysis}
\end{table*}

Tab.~\ref{tab:analysis} is computed from the episodes of the main table.
Most constraints are root waypoints, which decide where the body walks and stops and how it goes around furniture.
More than half of the segments of Opus 5.5 (54\%) are committed only after one or more revisions, 2.4 tries per segment on average, and Opus 5.5 issues a measurement after 93\% of its tries, for example where the hand stopped relative to the object.
Its revisions edit the constraints that shape the motion: only 6\% of them change nothing but the seed. InteractAgent$^\dagger$ can only reword its text prompts: it misses 83\% of its grasps, and after a miss its next try succeeds in 4\% of the cases.

All five language models measure after most of their tries (80\% to 99\%), so the weaker ones do not lack feedback; they differ in how they use it.
GLM-5.3-Flash and gpt-6-luna rarely write an end-effector target or a full-body keyframe (17\% and 4\% of their segments), miss three quarters of their grasps, and after a miss a revised try succeeds in 24\% of the cases, against 47\% for Claude Opus 5.5.
Gemini 4 Argon uses these constraints most often (49\%) and misses the fewest grasps.
The failed episodes of the weaker language models mostly use up the budget: 40 of 42 for GLM-5.3-Flash and 45 of 52 for gpt-6-luna.

\subsection{Spatial control (Sec.~\ref{sec:control_exp})}
\label{sec:supp_control}
On the ablation set, the two failures with root waypoints alone (Tab.~\ref{tab:control}) are reaches into the fridge and across a counter, where the palm stayed 0.1 to 0.2\,m short of the objects, and an end-effector target on a foot lets the agent tap a ball under a table where the kicks from text prompts passed over it.
Tab.~\ref{tab:supp_control_pertask} gives the result of every task of the ablation set for the five settings.

\begin{table}[h]
\centering
\scriptsize
\setlength{\tabcolsep}{4pt}
\begin{tabular}{@{}lccccc@{}}
\toprule
Task & Skills & Text & + root & + end eff. & + keyfr.\\
\midrule
low\_07 & $\cdot$ & $\cdot$ & \checkmark & \checkmark & \checkmark\\
low\_11 & $\cdot$ & $\cdot$ & \checkmark & \checkmark & \checkmark\\
low\_12 & $\cdot$ & $\cdot$ & \checkmark & \checkmark & \checkmark\\
low\_16 & \checkmark & \checkmark & \checkmark & \checkmark & \checkmark\\
low\_18 & $\cdot$ & $\cdot$ & \checkmark & \checkmark & \checkmark\\
low\_21 & $\cdot$ & $\cdot$ & \checkmark & $\cdot$ & \checkmark\\
low\_24 & $\cdot$ & \checkmark & \checkmark & \checkmark & \checkmark\\
low\_26 & \checkmark & $\cdot$ & \checkmark & \checkmark & \checkmark\\
low\_30 & \checkmark & \checkmark & \checkmark & \checkmark & \checkmark\\
mid\_06 & $\cdot$ & $\cdot$ & \checkmark & \checkmark & \checkmark\\
mid\_15 & $\cdot$ & $\cdot$ & \checkmark & \checkmark & \checkmark\\
mid\_17 & \checkmark & $\cdot$ & \checkmark & \checkmark & \checkmark\\
mid\_18 & $\cdot$ & $\cdot$ & \checkmark & \checkmark & \checkmark\\
high\_03 & $\cdot$ & $\cdot$ & $\cdot$ & \checkmark & \checkmark\\
high\_07 & \checkmark & $\cdot$ & \checkmark & \checkmark & \checkmark\\
high\_10 & $\cdot$ & $\cdot$ & $\cdot$ & \checkmark & \checkmark\\
posture\_01 & \checkmark & $\cdot$ & \checkmark & \checkmark & \checkmark\\
posture\_03 & $\cdot$ & \checkmark & \checkmark & \checkmark & $\cdot$\\
posture\_05 & $\cdot$ & $\cdot$ & \checkmark & \checkmark & \checkmark\\
posture\_09 & $\cdot$ & $\cdot$ & \checkmark & \checkmark & \checkmark\\
\midrule
\textbf{solved} & 6 & 4 & 18 & 19 & 19\\
\bottomrule
\end{tabular}
\caption{Per-task results of the settings of Tab.~\ref{tab:control} on the ablation set, in the order of its rows: atomic skills, text prompts only, and the three types of spatial constraints added one at a time. \checkmark\ marks a success.}
\label{tab:supp_control_pertask}
\end{table}

\subsection{Feedback (Sec.~\ref{sec:feedback_exp})}
\label{sec:supp_nocheck}
Without body measurements (Tab.~\ref{tab:feedback}) the agent estimates the hand from the joint rotations it wrote and, in one episode, needs 20 tries for a single grasp in the fridge. Without scene measurements it cannot follow a thrown object and must infer where it went from the final state.
Tab.~\ref{tab:supp_feedback_pertask} gives the result of every task of the ablation set for the six settings.

\begin{table}[h]
\centering
\scriptsize
\setlength{\tabcolsep}{4pt}
\begin{tabular}{@{}lcccccc@{}}
\toprule
Task & 1 & 2 & 3 & 4 & 5 & 6\\
\midrule
low\_07 & $\cdot$ & \checkmark & \checkmark & \checkmark & \checkmark & $\cdot$\\
low\_11 & \checkmark & \checkmark & \checkmark & \checkmark & \checkmark & \checkmark\\
low\_12 & \checkmark & \checkmark & \checkmark & \checkmark & \checkmark & \checkmark\\
low\_16 & \checkmark & \checkmark & \checkmark & \checkmark & \checkmark & \checkmark\\
low\_18 & \checkmark & \checkmark & \checkmark & \checkmark & \checkmark & \checkmark\\
low\_21 & \checkmark & \checkmark & \checkmark & \checkmark & \checkmark & $\cdot$\\
low\_24 & \checkmark & \checkmark & \checkmark & \checkmark & \checkmark & $\cdot$\\
low\_26 & \checkmark & \checkmark & \checkmark & \checkmark & \checkmark & \checkmark\\
low\_30 & \checkmark & \checkmark & \checkmark & \checkmark & \checkmark & \checkmark\\
mid\_06 & \checkmark & \checkmark & \checkmark & \checkmark & \checkmark & \checkmark\\
mid\_15 & $\cdot$ & \checkmark & \checkmark & \checkmark & $\cdot$ & \checkmark\\
mid\_17 & \checkmark & \checkmark & \checkmark & \checkmark & \checkmark & \checkmark\\
mid\_18 & \checkmark & \checkmark & \checkmark & \checkmark & \checkmark & \checkmark\\
high\_03 & $\cdot$ & \checkmark & \checkmark & \checkmark & \checkmark & $\cdot$\\
high\_07 & \checkmark & $\cdot$ & \checkmark & \checkmark & $\cdot$ & $\cdot$\\
high\_10 & \checkmark & \checkmark & \checkmark & \checkmark & \checkmark & \checkmark\\
posture\_01 & \checkmark & \checkmark & \checkmark & \checkmark & \checkmark & \checkmark\\
posture\_03 & \checkmark & \checkmark & \checkmark & $\cdot$ & \checkmark & $\cdot$\\
posture\_05 & \checkmark & \checkmark & $\cdot$ & \checkmark & $\cdot$ & \checkmark\\
posture\_09 & \checkmark & \checkmark & \checkmark & \checkmark & \checkmark & \checkmark\\
\midrule
\textbf{solved} & 17 & 19 & 19 & 19 & 17 & 14\\
\bottomrule
\end{tabular}
\caption{Per-task results of the six settings of Tab.~\ref{tab:feedback} on the ablation set, numbered as its rows. \checkmark\ marks a success.}
\label{tab:supp_feedback_pertask}
\end{table}

\begin{table}[h]
\centering
\small
\setlength{\tabcolsep}{3pt}
\resizebox{\linewidth}{!}{%
\begin{tabular}{lcccccccccc}
\toprule
Success check & All & low & mid & high & posture & Sub-goal & Collisions & Disturbed & Calls & Tokens\\
\midrule
with & 98.7 & 100.0 & 100.0 & 100.0 & 90.0 & 99.6 & 9.3 & 1.0 & 51 & 99k\\
without & 86.7 & 96.7 & 85.0 & 80.0 & 70.0 & 92.0 & 8.2 & 0.7 & 58 & 128k\\
\bottomrule
\end{tabular}}
\caption{Inhabitant with Claude Opus 5.5 and all measurements on the 75 tasks of Inhabit, with and without the success check. Success rates and sub-goals in \%, the other columns are means per episode as in Tab.~\ref{tab:main}. Without the check the agent ends the episode itself, and the episode is judged on its final state.}
\label{tab:supp_nocheck}
\end{table}

Without the check the agent fails on ten tasks. In seven of them it declared the task done although the success rule was not satisfied, most of them throws and final postures, and in the other three it used up the budget. Of the 72 episodes that it ended as done, 65 are successes.
On the ablation set (Tab.~\ref{tab:feedback}), the two extra failures with all measurements are tasks the agent declared done: it sat on the front edge of a chair, and it carried a ball to the target by hand where the ball had to be kicked. With the rendered frames only, three of the six failures are again tasks declared done: a posture that does not count as lying, a toss released closer than the required metre, and a three-part tidy-up with one part not done.

\subsection{The loop with other control specifications (Sec.~\ref{sec:loop_exp})}
\label{sec:supp_loop_settings}
\begin{table}[h]
\centering
\small
\resizebox{\linewidth}{!}{%
\begin{tabular}{lccc}
\toprule
Control specification & Full loop & Single-Try & Open-Loop\\
\midrule
Full (Inhabitant) & \textbf{19} (7.8) & 15 (30.4) & 0 (10.9)\\
Text prompts only & 4 (7.2) & 8 (178.3) & 0 (23.2)\\
HumanCLAW skills & 6 (8.4) & 6 (23.9) & 1 (20.4)\\
\midrule
Full, no measurements & 17 (14.6) & \multicolumn{2}{c}{}\\
\bottomrule
\end{tabular}}
\caption{All ten settings on the ablation set with Claude Opus 5.5: tasks solved out of 20 (mean collisions per task).}
\label{tab:ablation_full}
\end{table}
With text prompts only, Single-Try solves 8 tasks and the full loop 4: without the option to revise, the agent commits imprecise segments and corrects them with further segments, at the cost of 178 collisions per task, while with the full loop it spends the budget on retrying seeds.
With HumanCLAW's skills both solve 6 tasks.
\begin{figure}[h]
\centering
\includegraphics[width=0.95\linewidth]{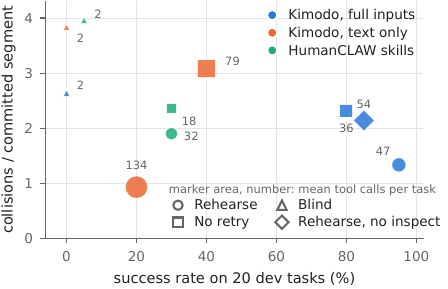}
\caption{Success against collisions per committed segment for the ten settings; marker area is the mean number of tool calls per task.}
\label{fig:tradeoff}
\end{figure}

\subsection{Tracking the motions with a physics controller (Sec.~\ref{sec:loop_exp})}
\label{sec:supp_phc}

Half-physics has no balance model (Sec.~\ref{sec:exec}), so we test whether a physically simulated humanoid with balance and torques can follow the interaction sequences.
We track the interaction sequence of each of the 75 episodes of Claude Opus 5.5 with the pretrained PHC+ controller~\cite{luo2023perpetual} in Isaac Gym, on a flat floor without furniture or objects.
Parts where the body is supported by furniture cannot be tracked without the furniture, so we remove every interval in which a body part other than the feet touches furniture for at least 0.3\,s or a foot stands on it, with 0.5\,s on each side, and track each remaining window of at least 1.5\,s separately, starting from the pose and velocity of its first frame.
This gives 171 windows covering 73\% of the frames.
We leave out the first 0.5\,s of each window and count a window as failed when the mean joint error exceeds 0.5\,m, the criterion of PHC.

\begin{table}[h]
\centering
\small
\begin{tabular}{lc}
\toprule
Windows (frames covered) & 171 (73\%)\\
Tracked without failure & 161 (94\%)\\
Global MPJPE & 53.7\,mm\\
Root-relative MPJPE & 40.5\,mm\\
Acceleration error & 6.0\,mm/frame$^2$\\
\midrule
Global MPJPE, walking & 52.8\,mm\\
Global MPJPE, low postures & 110.8\,mm\\
Global MPJPE, other & 47.6\,mm\\
\bottomrule
\end{tabular}
\caption{PHC+ tracking of the interaction sequences of Claude Opus 5.5 on all 75 tasks, without furniture and objects. Low postures are frames with the head below 1.2\,m: bending, squatting, kneeling and lying.}
\label{tab:supp_phc}
\end{table}

The controller follows 161 of the 171 windows without falling, at a mean joint error of 54\,mm (Tab.~\ref{tab:supp_phc}).
The ten failures fall in ten different tasks.
Five of them are among the 13 windows of the tasks that end in a posture, where the body kneels or lies down, and low postures also have the largest error.
In the tasks without a posture only 5 of 158 windows fail.
The test has limits.
Without furniture and objects it leaves out the supported parts of the motion and the weight of carried objects, the reference is the body as executed in half-physics, and the numbers are not comparable to tracking results on motion capture data.

\section{Video}
\label{sec:supp_video}
The video on the project page compares Inhabitant and the three baselines, and all agents in it use the same language model.
It first summarises which of the two interfaces each agent has.
It then plays three tasks with the four agents side by side, so that the differences in the motion are visible: the baselines bump into furniture and walls, miss their grasps or stop, while Inhabitant completes the task.
Next it follows the loop of Sec.~\ref{sec:loop} on one task: the tries of a segment are shown one after another with the measurement that the agent requested and the change that it made to the control specification before the next try.
The video then shows one long-horizon task from start to end with the number of tries of every segment, and more tasks of the benchmark, including throwing, moving objects with a foot, drawers, the fridge and postures.

\end{document}